\documentclass{article}

\usepackage[utf8]{inputenc} 
\usepackage[T1]{fontenc}    
\usepackage{pifont}
\usepackage{fullpage}
\usepackage{multicol,lipsum,xcolor}
\usepackage{hyperref}       
\usepackage{url}            
\usepackage{amsmath,amsthm,amssymb,bm}
\usepackage{bbold}
\DeclareMathOperator{\argmin}{argmin}
\DeclareMathOperator{\argmax}{argmax}
\usepackage{algorithm,algorithmic}
\usepackage{caption}
\usepackage{graphicx}
\usepackage{enumerate}
\usepackage{enumitem}
\usepackage{natbib}
\usepackage{mathtools}

\newtheorem{thm}{Theorem}
\newtheorem{lemma}[thm]{Lemma}

\theoremstyle{definition}

\newtheorem{defini}{Definition}

\newtheorem{assumption}{Assumption}

\newenvironment{itemize*}%
{\begin{itemize}[leftmargin=*,topsep=0pt]%
		\setlength{\itemsep}{0pt}%
		\setlength{\parskip}{0pt}}%
	{\end{itemize}}
\newenvironment{enumerate*}%
{\begin{enumerate}[leftmargin=*,topsep=0pt]%
		\setlength{\itemsep}{0pt}%
		\setlength{\parskip}{0pt}}%
	{\end{enumerate}}

\newcount\Comments  
\definecolor{darkpink}{rgb}{0.92, 0.2, 0.45}
\definecolor{teal}{rgb}{0.3,0.8,0.8}
\definecolor{forestgreen}{rgb}{0.13, 0.55, 0.13}
\newcommand{\kibitz}[2]{\ifnum\Comments=1{\textcolor{#1}{\textsf{\footnotesize #2}}}\fi}
\newcommand{\ambuj}[1]{\kibitz{darkpink}{[AT: #1]}}

\newcommand*\diff{\mathop{}\!\mathrm{d}}

\newcommand\blfootnote[1]{%
  \begingroup
  \renewcommand\thefootnote{}\footnote{#1}%
  \addtocounter{footnote}{-1}%
  \endgroup
}

\title{Bandit Algorithms for Precision Medicine}
\author{Yangyi Lu$^*$, Ziping Xu$^*$, Ambuj Tewari \\
{\tt \{yylu,zipingxu,tewaria\}@umich.edu} \\
\ \\
Department of Statistics\\
University of Michigan
\blfootnote{These authors made equal contributions to the writing of this chapter.}}

\begin{document}

\maketitle

\section{Introduction}

The Oxford English Dictionary defines precision medicine as ``medical care designed to optimize efficiency or therapeutic benefit for particular groups of patients, especially by using genetic or molecular profiling.'' It is not an entirely new idea: physicians from ancient times have recognized that medical treatment needs to consider individual variations in patient characteristics \citep{konstantinidou2017origins}. However, the modern precision medicine movement has been enabled by a confluence of events: scientific advances in fields such as genetics and pharmacology, technological advances in mobile devices and wearable sensors, and methodological advances in computing and data sciences.

This chapter is about bandit algorithms: an area of data science of special relevance to precision medicine. With their roots in the seminal work of Bellman, Robbins, Lai and others, bandit algorithms have come to occupy a central place in modern data science (see the book by \cite{lattimore2020bandit} for an up-to-date treatment). Bandit algorithms can be used in any situation where treatment decisions need to be made to optimize some health outcome. Since precision medicine focuses on the use of patient characteristics to guide treatment, \emph{contextual} bandit algorithms are especially useful since they are designed to take such information into account.

The role of bandit algorithms in areas of precision medicine such as mobile health and digital phenotyping has been reviewed before \citep{tewari17ads,rabbi19optimizing}. Since these reviews were published, bandit algorithms have continued to find uses in mobile health and several new topics have emerged in the research on bandit algorithms.
This chapter is written for quantitative researchers in fields such as statistics, machine learning, and operations research who might be interested in knowing more about the algorithmic and mathematical details of bandit algorithms that have been used in mobile health.

We have organized this chapter to meet two goals. First, we want to provide a concise exposition of basic topics in bandit algorithms. Section~\ref{sec:basics} will help the reader become familiar with basic problem setups and algorithms that appear frequently in applied work in precision medicine and mobile health (see, for example, \cite{paredes2014poptherapy,piette2015potential,rabbi2015mybehavior,piette2016patient,yom2017encouraging,rindtorff2019biologically,forman2019can,liao2020personalized,ameko2020offline,aguilera2020mhealth,tomkins2021intelligentpooling}). Second, we want to highlight a few advanced topics that are important for mobile health and precision medicine applications but whose full potential remains to be realized. Section~\ref{sec:advs} will provide the reader with helpful entry points into the bandit literature on non-stationarity, robustness to corrupted rewards, satisfying additional constraints, algorithmic fairness, and causality.



\section{Basic Topics} \label{sec:basics}

In this section, we begin by introducing the most simple of all bandit problems: the multi-armed bandit. Then we discuss a more advanced variant called contextual bandit that is especially suitable for precision medicine applications. The last topic we discuss in this section is offline learning which deals with algorithms that can use already collected data. The offline learning setting is to be contrasted with the online learning setting where the bandit algorithm has control over the data it collects.

\subsection{Multi-armed Bandit}\label{sec:MAB}
In recent years, the multi-armed bandit (MAB) framework has attracted a lot of attention in many application areas such as healthcare, marketing, and recommendation systems. 
MAB is a simple model that describes the interaction between an \emph{agent}\footnote{Also referred to as a \emph{learner}, \emph{statistician}, or \emph{decision maker}.} and an \emph{environment}.
At every time step, the agent makes a choice from an action\footnote{Since the historical roots of probability theory lie in gambling and casinos, it is not surprising that the MAB terminology comes from imagining a slot machine in a casino. A slot machine is also called a ``one-armed bandit'' as it robs you of your money. Therefore, we will use \emph{actions} and \emph{arms} interchangeably.} set and receives a reward. 
The agent may have different goals, such as maximizing the (discounted) cumulative reward within a time horizon, identifying the best arm, or competing with the arm with the best risk-return trade-off etc. 
In this section, we focus on maximizing the cumulative rewards for simplicity.
An important observation is that the agent needs to balance between \emph{exploration} and \emph{exploitation} to achieve its goal of receiving high cumulative reward. 
That is, both under-explored arms as well as tested-and-tried arms with high rewards should be selected often but for different reasons: the former have the potential to achieve high rewards and the latter are already confirmed to be good based on the past experience. 

To formally define the bandit framework, we start with introducing some notation. 
Suppose the agent interacts with the environment for $T$ time steps, where $T$ is called the horizon. 
In each round $t\in[T]$, the learner chooses an action $A_t$ from the action set $\mathcal{A}$ and receives a corresponding reward $R_t\in\mathbb{R}$.
We denote the cardinality of $\mathcal{A}$ by $K$.
The choice of $A_t$ depends on the action/reward history up to time $t-1$: $H_{t-1} = \left(A_1,R_1,\ldots,A_{t-1},R_{t-1}\right)$. 
A policy $\pi_t$ is defined as a mapping from the history up to time $t-1$ to the actions.
For short, we use $\pi$ as the sequence of policies $(\pi_0, \ldots, \pi_{T-1})$. 

In a healthcare setting, the fundamental pattern that often occurs is the following. Of course, this simple pattern fails to capture the full complexity of decision making in healthcare, but it is a reasonable starting point, especially for theoretical analysis. 

\begin{algorithm}
	\caption{Bandit Framework in Healthcare}
	\centering
	\begin{algorithmic}
		\STATE \textbf{Input:} Available treatment options, treatment period length $T$.
		\FOR{$t=1,\ldots,T$}
		\STATE A treatment option (action) is selected and delivered to the patient.
		\STATE Patient's health outcome (reward) following the treatment is recorded.
		\ENDFOR
	\end{algorithmic}
\end{algorithm}

In the remainder of this section, we will review bandit algorithms that learn good decision policies over time. 
We focus on the two key settings: \emph{stochastic bandit} and \emph{adversarial bandit}.
In both settings, the algorithms aim at minimizing their regret, which measures the difference between the maximal reward one can get, and the reward obtained by the algorithm. 
We will formally define regret in each setting. 

\subsubsection{Stochastic Multi-armed Bandit}
A stochastic bandit is a set of distributions $\nu = (P_a: a\in\mathcal{A})$ and we define the environment class $\mathcal{E}$ as a set of such distributions
\begin{align*}
    \mathcal{E} = \{\nu = (P_a: a\in\mathcal{A}):P_a\in\mathcal{M}_a \text{ for all }a\in\mathcal{A}\},
\end{align*}
where for each $a$, $\mathcal{M}_a$ is a set of distributions. 
For unstructured bandits, playing one action cannot help the agent deduce anything about other actions. 
Environment classes that are not unstructured are called structured, such as linear bandits~\citep{abbasi2011improved}, low-rank bandits~\citep{lu2021low} and combinatorial bandits~\citep{cesa2012combinatorial} etc. 
Throughout this chapter, we assume all bandit instances are $\mathcal{E}_{\mathrm{SG}}^K(1)$, which means the reward distribution for all $K$ arms is 1-subgaussian.
\begin{defini}[Subgaussianity]
A random variable $X$ is $\sigma$-subgaussian if for all $\lambda\in\mathbb{R}$, $\mathbb{E}\left[e^{\lambda X}\right]\leq e^{\lambda^2\sigma^2/2}$. 
\end{defini}
It is not hard to see from the definition that many well-known distributions are subgaussian, e.g., any bounded-domain distribution, Bernoulli distribution and Gaussian distributions.
Intuitively, a subgaussian distribution has tails no heavier than a Gaussian distribution.
Many nice concentration inequalities have been developed for subgaussian variables and are widely used in the proofs of bandit algorithms.

In the process of interactions, once the agent performs action $a_t$ following a particular policy, the environment samples a reward $R_t$ from the distribution $P_{a_t}$.
The combination of an environment and agent policy induces a probability measure on the sequence of outcomes $a_1, R_1, \ldots, a_T, R_T$. 
A standard stochastic MAB protocol is following. At every time step $t = 1,\ldots, T$, the learning agent
\begin{enumerate}
	\item picks an action $a_t\in\mathcal{A}$ following policy $\pi_{t-1}$,
	\item receives reward $R_t\sim P_{a_t}$,
	\item updates its policy to $\pi_t$.
\end{enumerate}
We note that $P_{a_t}$ is the conditional reward distribution of $R_t$ given $\{H_{t-1}, a_t\}$ and $\pi_t$ is a function from $H_{t-1}$ to $\mathcal{A}$.
The expected reward of action $a$ is defined by $\mu_a(\nu) \overset{\text{def}}{=}\mathbb{E}_\nu[R|a] = \int_{-\infty}^{\infty}r d P_a(r)$, where $R$ is used as the reward variable.
Then the maximum expected reward and the optimal action are given by
\begin{align*}
	\mu^*(\nu) = \max_{a\in\mathcal{A}} \mu_a(\nu) \text{ and } a^*(\nu) \in \argmax_{a\in\mathcal{A}} \mu_a(\nu).
\end{align*}
According to above definition, more than one optimal actions can exist and the optimal policy is to select an optimal action at every round.
For actions whose expected rewards are less than optimal actions, we call them sub-optimal actions and define the reward gap between action $a$ and $a^*(\nu)$ by $\Delta_a:=\mu^*(\nu) - \mu_a(\nu)$.

As mentioned earlier, the learner's goal is to maximize the cumulative reward $S_T = \sum_{t=1}^{T}R_t$. 
We now define a performance metric called \emph{regret} which is the difference between the expected reward that $\pi^*(\nu)$ can obtain and $\mathbb{E}[S_T]$.
Minimizing the regret is equivalent to maximizing the reward.
The reason why we do not directly optimize $S_T$ is that the cumulative rewards depends on the environment and it is hard to tell whether a policy is good or not by merely looking at the cumulative rewards unless it is compared to a good policy.
So we define the problem-dependent regret of a policy $\pi$ on bandit instance $\nu$ by
\begin{align*}
	\text{Reg}_T(\pi, \nu) = T\mu^*(\nu) - \mathbb{E}[S_T], 
\end{align*}
where the expectation is taken over actions and rewards up to time $T$.  
The worst-case regret of a policy $\pi$ is defined by
\begin{align*}
	\text{Reg}_T(\pi) = \sup_{\nu\in\mathcal{E}} \text{Reg}_T(\pi,\nu).
\end{align*}
We will drop $\pi$ and $\nu$ from the regret when they are clear from the context.

\paragraph{Remark about Pure Exploration. }
Throughout this chapter, we focus on minimizing regret by balancing exploration and exploitation. 
We also want to point out that in a different setting, the exploration cost may not be a concern and the agent just wants to output a final recommendation for the best arm after an exploration phase. 
Problems of this type are called pure exploration problems.
In such cases, algorithms are usually evaluated by sample complexity or simple regret~\citep{bubeck2009pure}.
Pure exploration is also related to randomized controlled trials (RCTs) including modern variants that involve sequential randomization such as sequential multiple assignment randomized trials (SMARTs) \citep{lei2012smart} and micro-randomized trials (MRTs) \citep{klasnja2015microrandomized}. Randomized trials are typically designed to enable estimation of treatment effects with sufficient statistical power. Since the concerns of pure exploration and randomized controlled trials are different from those of bandit algorithms, we do not discuss them further in this chapter. However, note that in an actual application, methodology from bandits and randomized trials may need to be integrated. Researchers may start off with a randomized trial and follow it up with bandit algorithm in the next iteration of their health app. They can also decide to run a randomized trial for one health outcome while simultaneously running a bandit for a different outcome (e.g., an outcome related to user engagement with the health app) in the same study. There is also ongoing work \citep{yao2020power,zhang2021statistical} on enabling the kind of statistical analysis done after randomized trials on data collected via online bandit algorithms.

\paragraph{Explore-then-Commit (ETC).}
We start with a simple two-stage algorithm: Explore-then-Commit (ETC). 
In the first stage of ETC, the learner plays every arm for a fixed number of times ($m$) and obtain estimates of the expected rewards. 
In the second stage, the learner commits to the best arm according to the estimates in the first stage. 
For every arm $a$, let $\hat{\mu}_a(t)$ denote the estimated expected reward up to time $t$:
\begin{align*}
	\hat{\mu}_a(t) = \frac{1}{T_a(t)}\sum_{s=1}^{t} \mathbb{1}_{\{A_s = a\}} R_s,
\end{align*}
where $T_a(t) = \sum_{s = 1}^{t}\mathbb{1}_{\{A_s = a\}}$ is the number of times action $a$ has been performed up to round $t$.

\begin{algorithm}[t]
	\caption{Explore-then-Commit (ETC)} 
	\centering
	\begin{algorithmic}[1]
		\STATE \textbf{Input:} action space $\mathcal{A}$, where $K = |\mathcal{A}|$, number of exploration steps, $m$, horizon $T$.
		\FOR{$t=1,\ldots,T$}
		\IF{$t \leq mK$}
		\STATE play $A_t = (t \text{ mod } K) + 1$.
		\ELSE
		\STATE play $A_t = \argmax_a \hat{\mu}_a(mK)$.
		\ENDIF
		\STATE receive $R_t$.
		\ENDFOR
	\end{algorithmic}
\label{algo:ETC}
\end{algorithm}

With the above definitions, we are ready to present the ETC in Algorithm~\ref{algo:ETC}. 
The overall performance of ETC crucially depends on the parameter $m$. 
If $m$ is too small, the algorithm cannot estimate the performance of every arm accurately, so it is likely to exploit a sub-optimal arm in the second stage, which leads to high regret. 
If $m$ is too big, the first stage (explore step) plays with sub-optimal arms for too many times, so that the regret can be large again. 
The art is to choose an optimal value for $m$ in order to minimize the total regret incurred in both stages. 
Specifically, ETC achieves $O(T^{2/3})$ worst-case regret\footnote{Ignoring parameters other than $T$.} by choosing $m = O(T^{2/3})$~\citep{lattimore2020bandit}.
Sub-linear regret $O(T^{2/3})$ performance is good as a starting point. 
Next, we will introduce another two classic algorithms which incur even less regret.

\paragraph{Upper Confidence Bound (UCB).}
There are several types of exploration strategies to select actions such as greedy, Boltzmann, optimism and pessimism. 
Suppose the agent has reward estimates $\hat{\mu}_a$ for all actions.
A greedy exploration strategy simply selects the action with the highest $\hat{\mu}_a$.
A Boltzmann exploration strategy picks each action with probability proportional to $\exp(\eta \hat{\mu}_a)$, where $\eta$ is a tuning parameter.
Boltzmann becomes greedy as $\eta$ goes to infinity. 
For an optimism strategy, one picks the action with the highest reward estimate plus some bonus term, i.e. $\argmax_{a}\hat{\mu}_a + \text{bonus}_a$.
In contrast, a pessimism strategy would pick the action: $\argmax_{a}\hat{\mu}_a - \text{bonus}_a$.

Out of these strategies, UCB algorithm follows the optimism strategy, in particular, a famous principle called \emph{optimism in the face of uncertainty (OFU)}, which means that one should act as if the environment is the best possible one among those that are \emph{plausible} given current experience. 
The reason OFU works is that misplaced optimism gets corrected when under-explored actions are tried and low rewards are observed.
In contrast, pessimism does not work (at least in the online setting; for the offline setting things can be different \citep{jin2021pessimism}) since wrong beliefs about low performance of under-explored actions do not get a chance to get revised by collecting more data from those actions.

At every step $t$, the UCB algorithm updates a value called \emph{upper confidence bound} defined for each action $a \in \mathcal{A}$ and confidence level $\delta \in [0, 1]$ as follows.
\begin{align}
	\text{UCB}_a(t-1,\delta) = \begin{cases}
	\infty, &\text{ if } T_a(t-1) = 0\\
	\hat{\mu}_a(t-1)+\sqrt{\frac{2\log(1/\delta)}{T_a(t-1)\vee 1}}, &\text{ otherwise}. \label{equ:UCB}
	\end{cases}
\end{align}
The learner chooses the action with the highest UCB value at each step.
Overall, UCB (Algorithm~\ref{algo:UCB}) guarantees a $\tilde{O}(\sqrt{T})$ worst-case regret (where the informal $\tilde{O}(\cdot)$ notation hides constants and logarithmic factors).

\begin{algorithm}[t]
	\centering
	\caption{Upper Confidence Bound (UCB)}
	\begin{algorithmic}[1]
		\STATE \textbf{Input: } $\delta$, $\mathcal{A}$ where $K = |\mathcal{A}|$. 
		\FOR{$t = 1,\ldots, T$}
		\STATE play $A_t = \argmax_{a \in \mathcal{A}} \text{UCB}_a(t-1,\delta)$.
		\STATE receive reward $R_t$ and update the upper confidence bound terms according to \eqref{equ:UCB}.
		\ENDFOR
	\end{algorithmic}
\label{algo:UCB}
\end{algorithm}

According to the construction of upper confidence bounds, an action will be selected under two circumstances: under-explored ($T_a(t-1)$ small) or well-explored with good performance ($\hat{\mu}_a(t-1)$ large).
The upper confidence bound for an action gets close to its true mean after being selected for enough times. 
A sub-optimal action will only be played if its upper confidence bound is larger than that of the optimal arm. 
However, this is unlikely to happen too often. 
The upper confidence bound for the sub-optimal action will eventually fall below that of the optimal action as we play the sub-optimal actions more times. 
We present the regret guarantee for UCB (Algorithm~\ref{algo:UCB}) in Theorem~\ref{thm:UCB}.

\begin{thm}[Regret for UCB Algorithm]
	\label{thm:UCB}
	If $\delta = 1/T$, then the problem-dependent regret of UCB, as defined in Algorithm~\ref{algo:UCB}, on any 1-subgaussian bandit $\nu$ is bounded by
	\begin{align}
		\mathrm{Reg}_T(\mathrm{UCB},\nu)  \leq 3\sum_{a\in\mathcal{A}}\Delta_a + \sum_{a:\Delta_a>0} \frac{16\log T}{\Delta_a}, 
	\end{align}
	where $\Delta_a := \mu^*(\nu) - \mu_a(\nu)$ represents the corresponding gap term. 
	The worst-case regret bound of UCB is:
	\begin{align}
	    \mathrm{Reg}_T(\mathrm{UCB}) = O(\sqrt{KT\log T}).
	\end{align}
\end{thm}
\begin{proof}
    We only present the worst-case regret for simplicity. For the problem-dependent regret proof, we refer the reader to Chapter 7 in \cite{lattimore2020bandit}.
    
    We define a good event $E$ as follows:
    \begin{align*}
        E:=\left\{|\hat{\mu}_a(t-1) - \mu_a| \leq \sqrt{\frac{2\log(1/\delta)}{T_a(t-1)\vee 1}},\forall t\in[T], a\in\mathcal{A}\right\}.
    \end{align*}
    By Hoeffding inequality and union bound, one can show that $\mathbb{P}(E^c)\leq 2TK\delta^4$. Next, we decompose the regret.
    \begin{align*}
        \mathrm{Reg}_T(\mathrm{UCB}) 
        & = \mathbb{E}\left[\sum_{t=1}^T\mu^*-\mathrm{UCB}_{a^*}(t-1,\delta)+\mathrm{UCB}_{a^*}(t-1,\delta)-\mathrm{UCB}_{A_t}(t-1,\delta)+\mathrm{UCB}_{A_t}(t-1,\delta)-\mu_{A_t}\right]\\
        & \leq \mathbb{E}\left[\sum_{t=1}^T\mu^*-\mathrm{UCB}_{a^*}(t-1,\delta)+\mathrm{UCB}_{A_t}(t-1,\delta)-\mu_{A_t}\right].
    \end{align*}
    The inequality in above expression is due to the action selection criterion in UCB algorithm. 
    Condition on event $E$, we have $\mu^*-\text{UCB}_{a^*}(t-1,\delta)\leq 0$ for all $t\in[T]$; otherwise, the regret can be bounded by $2T$. 
    Combining these arguments, we have
    \begin{align*}
        \mathrm{Reg}_T(\mathrm{UCB}) & \leq \mathbb{P}(E^c)\cdot 2T + \mathbb{P}(E)\cdot \mathbb{E}\left[\sum_{t=1}^T\mathrm{UCB}_{A_t}(t-1,\delta)-\mu_{A_t}\mid E\right]\\
        &\leq 4T^2K\delta^4 + 2\mathbb{E}\left[\sum_{t=1}^T\sqrt{\frac{2\log\left(1/\delta\right)}{T_{A_t}(t-1)\vee 1}}\right] \text{ (By definition of event $E$)}\\
        & \leq 4T^2K\delta^4 + \sqrt{8\log\left(1/\delta\right)}\sum_{a\in\mathcal{A}}\sum_{t=1}^T\mathbb{E}\left[\sqrt{\frac{1}{T_{A_t}(t-1)\vee 1}}\mathbb{1}_{\{A_t=a\}}\right]\\
        & \leq 4T^2K\delta^4 + \sqrt{8\log\left(1/\delta\right)}\sum_{a\in\mathcal{A}}\int_{1}^{T_a(T)}\sqrt{1/s} \diff s\\
        & \leq 4T^2K\delta^4 + \sqrt{8\log\left(1/\delta\right)}\sqrt{KT} = O(\sqrt{KT\log T}) \text{ (Set $\delta = 1/T$)}.
    \end{align*}
    The last inequality in above is by Cauchy-Schwarz inequality. 
\end{proof}

The UCB family has many variants, one of which is to replace the upper confidence bound for every action by 
	$\hat{\mu}_a(t-1)+\sqrt{{2\log\left(1+t\log^2(t)\right)}/{T_a(t-1)}}$.
Even though the regret dominant terms ($\sqrt{KT\log T}$ and $\sum_{a:\Delta_a>0} \frac{\log T}{\Delta_a}$) for this version has the same order as those of Algorithm~\ref{algo:UCB}, the leading constants for the two dominant terms become smaller. 

Then one may ask the question: 
is it possible to further improve the regret bound of UCB and above variant?
The answer is yes. 
\citet{audibert2009minimax} proposed an algorithm called \emph{MOSS} (Minimax Optimal Strategy in the Stochastic case). 
\emph{MOSS} replaces the upper confidence bounds in Algorithm~\ref{algo:UCB} by
\begin{align*}
	\hat{\mu}_a(t-1)+\sqrt{\frac{\max\left\{\log\left(\frac{T}{KT_a(t-1)}\right),0\right\}}{T_a(t-1)}}.
\end{align*}
Under this construction, the worst-case regret of \emph{MOSS} is guaranteed to be only $O(\sqrt{KT})$.

However, \emph{MOSS} is not always good.
One can easily construct regimes where the problem-dependent regret of MOSS is worse than UCB~\citep{lattimore2015optimally}. 
On the other hand, the improved UCB algorithm proposed by~\cite{auer2010ucb} satisfies a problem-dependent regret that is similar to \eqref{equ:UCB}, but the worst-case regret is $O(\sqrt{KT\log K})$. 
Later on, by carefully constructing the upper confidence bounds, Optimally Confidence UCB algorithm~\citep{lattimore2015optimally} and AdaUCB algorithm~\citep{lattimore2018refining} are shown to achieve $O(\sqrt{KT})$ worst-case regret and their problem-dependent regret bounds are also not worse than that of the UCB algorithm. 
There are many more UCB variants in the literature that we do not cover in this chapter. 
The reader may refer to Table 2 in \cite{lattimore2018refining} for a comprehensive summary.

\paragraph{Successive Elimination (SE). }
We now describe the SE algorithm that also relies on the upper confidence bound calculations. 
The idea is similar to UCB such that a sub-optimal arm is very unlikely to have large a upper confidence bound if it has been selected for enough times.
At every round, SE maintains a confidence interval for the mean reward of every arm and removes all arms whose reward upper bound is smaller than the lower bound of the biggest estimated reward arm. 
The procedure ends when there is only one arm remained.
We describe the SE algorithm in Algorithm~\ref{algo:SE} and define the UCB terms as \eqref{equ:UCB} and LCB terms as

\begin{align}
	\text{LCB}_a(t-1,\delta) = \begin{cases}
	\infty, &\text{ if } T_a(t-1) = 0\\
	\hat{\mu}_a(t-1)-\sqrt{\frac{2\log(1/\delta)}{T_a(t-1)\vee 1}}, &\text{ otherwise}. \label{equ:LCB}
	\end{cases}
\end{align}

\begin{algorithm}[t]
    \centering
    \caption{Successive Elimination (SE)}
    \begin{algorithmic}[1]
        \STATE \textbf{Input:} $\delta, \mathcal{A}$ where $K = |\mathcal{A}|$.
        \WHILE{$\mathcal{A}$ contains more than one arm}
        \STATE play every arm in $\mathcal{A}$ once and update the UCBs and LCBs using \eqref{equ:UCB} and \eqref{equ:LCB}.
        \STATE Eliminate all arms $a$ s.t. $\exists a'\in\mathcal{A}$ with $\text{UCB}_a(t-1,\delta)<\text{LCB}_{a'}(t-1,\delta)$.
        \ENDWHILE
    \end{algorithmic}
    \label{algo:SE}
\end{algorithm}
SE was first proposed in \cite{even2006action} along with a similar action elimination based algorithm: Median Elimination (ME).
They studied the probably approximately correct (PAC) setting \citep{haussler2018probably}.
In particular, \cite{even2006action} shows that for given $K$ arms, it suffices to pull the arms for $O(\frac{K}{\varepsilon^2}\log(1/\delta)$ times to find an $\varepsilon$-optimal arm with probability at least $1-\delta$.
It is not hard to prove that SE also satisfies the following regret bound.
\begin{thm}[Regret for SE Algorithm]
    If $\delta = 1/T$, the worst-case regret of SE over 1-subgaussian bandit environments is bounded by
    \begin{align}
        \mathrm{Reg}_T(\mathrm{SE}) = O(\sqrt{KT\log T}).
    \end{align}
\end{thm}
\begin{proof}
    Without loss of generality, we assume the optimal arm $a^*$ is unique.
    Define the event $E$ by $\{|\hat{\mu}_a(t) - \mu_a|\leq c_a(t,\delta), \forall a, t\}$, where $c_a(t,\delta) = \sqrt{\frac{2\log(1/\delta)}{T_a(t)\vee 1}}$ denotes the confidence set width.
    By Hoeffding inequality and union bound, one can show that $\mathbb{P}(E^c)\leq 2\delta^4TK$.
    
    Define $t$ as the last round when arm $a$ is not eliminated yet. According to the elimination criterion in SE, the reward gap term can be bounded as:
    \begin{align*}
        \Delta_a:=\mu^* - \mu_a \leq 2(c_{a^*}(t,\delta)+c_a(t,\delta)) = O(c_a(t,\delta)).
    \end{align*}
    The last equality holds as $T_a(t)$ and $T_{a^*}(t)$ differ at most by $1$ by construction. 
    Since $t$ is the last round $a$ being played, we have $T_a(t) = T_a(T)$ and $c_a(t) = c_a(T)$, which implies below property for all non-optimal arms $a$:
    \begin{align*}
        \Delta_a \leq O\left(\sqrt{\frac{\log(1/\delta)}{T_a(T)}}\right).
    \end{align*}
    We thus obtain that under event $E$,
    \begin{align*}
        \sum_{t=1}^T\mathbb{E}[\mu^*-\mu_{a_t}|E] &\leq \sum_{a\in\mathcal{A}\setminus\{a^*\}}T_a(T)\Delta_a\leq O(\sqrt{\log (1/\delta)})\sum_{a\in\mathcal{A}} \sqrt{T_a(T)} \leq O(\sqrt{KT\log (1/\delta)}),
    \end{align*}
    where the last inequality is by Cauchy-Schwarz inequality.
    Take $\delta = 1/T$, $\mathrm{Reg}_T(\mathrm{SE}) = O(\sqrt{KT\log T})$ by conditional expectation calculations.
\end{proof}

\paragraph{Thompson Sampling (TS). }
All of the methods we have mentioned so far select their actions based on a frequentist view.
TS uses one of the oldest heuristic~\citep{thompson1933likelihood} for choosing actions and addresses the exploration-exploitation dilemma based on a Bayesian philosophy of learning. 
The idea is simple.
Before the game starts, the agent chooses a prior distribution over a set of possible bandit environments. 
At every round, the agent samples an environment from the posterior and acts according to the optimal action in that environment. 
The exploration in TS comes from the randomization over bandit environments. 
At the beginning, the posterior is usually poorly concentrated, then the policy will likely explore.
As more data being collected, the posterior tends to concentrate towards the true environment and the rate of exploration decreases. 
We present the TS algorithm in Algorithm~\ref{algo:TS}. 

To formally describe how TS works, we start with several definition related to Bayesian bandits.
\begin{defini}[K-armed Bayesian bandit environment]
	A K-armed Bayesian bandit environment is a tuple $(\mathcal{E}, \mathcal{G}, Q, P)$ where $(\mathcal{E}, \mathcal{G})$ is a measurable space and $Q$ is a probability measure on $(\mathcal{E}, \mathcal{G})$ called the prior. $P = (P_{\nu a}: \nu\in\mathcal{E}, a\in\mathcal{A})$ is the reward distribution for arms in bandit $\nu$, where $|\mathcal{A}| = K$. 
\end{defini}
Given a K-armed Bayesian bandit environment $(\mathcal{E}, \mathcal{G}, Q, P)$ and a policy $\pi$, the Bayesian regret is defined as:
\begin{align*}
	\mathrm{BReg}_T(\pi, Q) = \int_\mathcal{E} \text{Reg}_T(\pi, \nu) dQ(\nu).
\end{align*}

\begin{algorithm}[t]
	\caption{Thompson Sampling (TS)}
	\centering
	\begin{algorithmic}[1]
		\STATE \textbf{Input: }Bayesian bandit environment $(\mathcal{E}, \mathcal{B}(\mathcal{E}), Q, P)$ ($\mathcal{B}(\cdot)$ is the Borel set), action set $\mathcal{A}$ with $|\mathcal{A}| = K$.
		\FOR{$t = 1,\ldots, T$}
		\STATE Sample $\nu_t\sim Q(\cdot|A_1, R_1, \ldots, A_{t-1}, R_{t-1})$.
		\STATE Play $A_t = \argmax_{i\in[K]} \mu_i(\nu_t)$.
		\ENDFOR
	\end{algorithmic}
\label{algo:TS}
\end{algorithm}
TS has been analyzed in both of the frequentist and the Bayesian settings and we will start with the Bayesian results.
\begin{thm}[Bayesian Regret for TS Algorithm]
	For a K-armed Bayesian bandit environment $(\mathcal{E}, \mathcal{G}, Q, P)$ such that $P_{\nu a}$ is 1-subgaussian for all $\nu\in\mathcal{E}$ and $a\in[K]$ with mean in $[0,1]$. Then the policy $\pi$ of TS satisfies
	\begin{align}
		\mathrm{BReg}_T(\pi, Q) = O(\sqrt{KT\log T}).
	\end{align}
\end{thm}
\begin{proof}
	The proof is quite similar to that of UCB. 
	We abbreviate $\mu_a = \mu_a(\nu)$ and let $a^* = \argmax_{a\in[K]}\mu_a$ be the optimal arm. 
	Note that $a^*$ is a random variable depending on $\nu$. 
	For every $a\in[K]$, we define a clipped upper bound term
	\begin{align*}
		\mathrm{UCB}_a(t-1) = \hat{\mu}_a(t-1)+\sqrt{\frac{2\log(1/\delta)}{1\vee T_a(t-1)}},
	\end{align*}
	where $\hat{\mu}_a(t-1)$ and $T_a(t-1)$ are defined in the same way as those in UCB.
	We define event $E$ such that for all $t\in[T]$ and $a\in\mathcal{A}$, 
	\begin{align*}
		|\hat{\mu}_a(t-1)-\mu_a| < \sqrt{\frac{2\log(1/\delta)}{1\vee T_a(t-1)}}. 
	\end{align*}
	By Hoeffding inequality and union bound, one can show that $\mathbb{P}(E^c) \leq 2TK\delta^4$. This result will be used in later steps. 
	
	Let $\mathcal{F}_t = \sigma(A_1,R_1,\ldots, A_t, R_t)$ be the $\sigma$-algebra generated by the interaction sequence up to time $t$. 
	The key insight for the whole proof is to observe below property from the definition of TS:
	\begin{align}
		\mathbb{P}(a^*|\mathcal{F}_{t-1}) = \mathbb{P}(A_t|\mathcal{F}_{t-1}) \text{ a.s. }  \label{equ:TS}
	\end{align}
	Using above property and $\text{BReg}_T = \mathbb{E}\left[\sum_{t=1}^{T}(\mu_{a^*}-\mu_{A_t})\right] = \mathbb{E}\left[\sum_{t=1}^{T}\mathbb{E}\left[(\mu_{a^*}-\mu_{A_t}) \mid \mathcal{F}_{t-1}\right]\right]$, we have
	\begin{align*}
		\mathbb{E}\left[\mu_{a^*}-\mu_{A_t} \mid \mathcal{F}_{t-1}\right] = \mathbb{E}\left[\mu_{a^*}- \text{UCB}_{a^*}(t-1) + \text{UCB}_{A_t}(t-1) - \mu_{A_t} \mid \mathcal{F}_{t-1}\right],
	\end{align*}
	and thus
	\begin{align*}
		\mathrm{BReg}_T = \mathbb{E}\left[\sum_{t=1}^{T}(\mu_{a^*}-\mathrm{UCB}_{a^*}(t-1)) + \sum_{t=1}^{T}(\mathrm{UCB}_{A_t}(t-1)-\mu_{A_t})\right].
	\end{align*}
	Conditioning on the high-probability event $E$, the first sum is negative and the second sum is of the order $O(\sqrt{KT\log (1/\delta)})$, while $\mathrm{BReg}_T \leq 2T$ if conditioning on $E^c$.
	Take $\delta = 1/T$, one can verify that $\text{BReg}_T = O(\sqrt{KT\log T})$.

\end{proof}

Compared to the analysis for Bayesian regret, frequentist (worst-case) regret analysis for TS gets a lot more technical. 
The key reason behind this is that the worst-case regret does not have an expectation with respect to the prior and therefore the property in \eqref{equ:TS} cannot be used. 
Even though TS was well-known to be easy to implement and competitive with state of the art methods, it lacked worst-case regret analysis for a long time. 
Significant progress was made by \citet{agrawal2012analysis} and \citet{kaufmann2012thompson}.
In \citet{agrawal2012analysis}, the first logarithmic bound on the frequentist regret of TS was proven. 
\citet{kaufmann2012thompson} provided a bound that matches the asymptotic lower bound of \citet{lai1985asymptotically}. 
However, both of these bounds were problem-dependent. 
The first near-optimal worst-case regret $O(\sqrt{KT\log T})$ was proved by \cite{agrawal2013further} for Bernoulli bandits with Beta prior, where the reward is either zero or one. 
For TS that uses Gaussian prior, the same work proved a $O(\sqrt{KT\log K})$ worst-case regret.
\citet{jin2020mots} proposed a variant of TS called MOTS (Minimax Optimal TS) that achieves $O(\sqrt{KT})$ regret.

\subsubsection{Adversarial Multi-armed Bandit}
In stochastic bandit models, the rewards are assumed to be strictly i.i.d.\ given actions.
This assumption can be violated easily in practice. 
For example, the health feedback for a patient after certain treatments may vary slightly across times and the way it varies is usually unknown. 
In such scenarios, a best action that maximizes the total reward still exists, but algorithms designed for stochastic bandit environments are no longer guaranteed to work. As a more robust counterpart to the stochastic model, we study the adversarial bandit model in this section, in which the assumption that a single action is good in hindsight is retained but the rewards are allowed to be chosen adversarially.


The adversarial bandit environment is often called as the adversary. 
In an adversarial bandit problem, the adversary secretly chooses reward vectors $r:=\{r_t\}_{t=1}^T$ where $r_t\in [0,1]^K$ corresponds to the rewards over all actions at time $t$. 
In every round, the agent chooses a distribution over the actions $P_t$.
An action $A_t\in[K]$ is sampled from $P_t$ and the agent receives the reward $r_{tA_t}$.
A policy $\pi$ in this setting maps the history sequences to distributions over actions. 
We evaluate the performance of policy $\pi$ by the expected regret, which is the cumulative reward difference between the best fixed action and the agent's selections:
\begin{align}
	\text{Reg}_T(\pi, r) = \max_{a\in\mathcal{A}}\sum_{t=1}^{T}r_{ta} - \mathbb{E}\left[\sum_{t=1}^{T}r_{tA_t}\right]
\end{align}
The worst-case regret of policy $\pi$ is defined by
\begin{align}
	\text{Reg}_T(\pi) = \sup_{r\in[0,1]^{T\times K}} \text{Reg}_T(\pi, r).
\end{align}
It may not be very clear at the beginning that why we define the regret by comparing to the fixed best action instead of the best action at every round. 
In the later case, the regret should be $\text{Reg}'_T(\pi, x) = \mathbb{E}\left[\sum_{t=1}^{T}\max_{a\in\mathcal{A}}r_{ta}-\sum_{t=1}^{T}r_{tA_t}\right]$.
However, this definition provides the adversary too much power so that for any policy, 
one can show $\text{Reg}'_T(\pi, r)$ can be $\Omega(T)$ for certain reward vectors $r\in[0,1]^{K\times T}$. 

\paragraph{Remark on randomized policy:} In stochastic bandit models, the optimal action is deterministic and the optimal policy is simply to select the optimal action at every round. 
However, in adversarial bandit setting, the adversary has great power in designing the reward.
It may know the agent's policy and design the rewards accordingly, so that a deterministic policy can incur linear regret. 
For example, we consider there are two actions, whose reward is either $0$ or $1$ at any time. 
For a deterministic policy, the agent decides to choose an action at time $t$.
Then the adversary knows it and can set the reward of that action at time $t$ to be $0$ and the reward of the unselected action to be $1$. 
The cumulative regret will be $T$ after $T$ rounds.
However, one can improve the performance by a randomized policy, e.g., choosing either action with probability $0.5$, then the adversary cannot make you incur regret $1$ at every round by manipulating the reward values for both actions. 

\paragraph{Exponential-weight algorithm for Exploration and Exploitation (EXP3).}
We now study one of the most famous adversarial bandit algorithm called EXP3.
Before describing the algorithm, we define some related terms below.
In a randomized policy, the conditional probability of the action $a$ being played is denoted by
\begin{align*}
	P_{ta} = \mathbb{P}(A_t = a\mid A_1, R_1,\ldots, A_{t-1}, R_{t-1}).
\end{align*}
Assuming $P_{ta}>0$ almost surely for all policies, a natural way to define the importance-weighted estimator of $r_{ta}$ is
\begin{align}
	\hat{R}_{ta} = \frac{\mathbb{1}_{\{A_t = a\}}R_t}{P_{ta}}. \label{equ:exp3_est1}
\end{align}
Let $\mathbb{E}_t[\cdot] = \mathbb{E}[\cdot\mid A_1, R_1,\ldots, A_{t-1}, R_{t-1}]$. 
A simple calculation shows that $\hat{R}_{ta}$ is conditionally unbiased, i.e. $\mathbb{E}_t[\hat{R}_{ta}] = r_{ta}$.
However, the variance of estimator $\hat{R}_{ta}$ can be extremely large when $P_{ta}$ is small and $r_{ta}$ is non-zero. 
Let $A_{ta}:=\mathbb{1}_{\{A_t=a\}}$, then the variance of the estimator is:
\begin{align*}
	\mathbb{V}_t[\hat{R}_{ta}] = \mathbb{E}_t[\hat{R}_{ta}^2] - r_{ta}^2 = \mathbb{E}_t\left[\frac{A_{ta}r_{ta}^2}{P_{ta}^2}\right] - r_{ta}^2 = \frac{r_{ta}^2(1-P_{ta})}{P_{ta}}.
\end{align*}
An alternative estimator is:
\begin{align}
	\hat{R}_{ta} = 1-\frac{\mathbb{1}_{\{A_t = a\}}}{P_{ta}}(1-R_t). \label{equ:exp3_est2}
\end{align}
This estimator is still unbiased and its variance is
\begin{align*}
	\mathbb{V}_t[\hat{R}_{ta}] = y_{ta}^2\frac{1-P_{ta}}{P_{ta}},
\end{align*}
where we define $y_{ta} = 1-r_{ta}$.

The best choice of the estimator $\hat{R}_{ta}$ depends on the actual rewards. 
One should use \eqref{equ:exp3_est1} for small rewards and \eqref{equ:exp3_est2} for large rewards. 
So far we have learned how to construct reward estimators for given sampling distributions $P_{ta}$.
EXP3 algorithm provides a way to design the $P_{ta}$ terms. 
Let $\hat{S}_{t,a} = \sum_{s=1}^{t}\hat{R}_{sa}$ be the total estimated reward until the end of round $t$, where $\hat{R}_{si}$ is defined in \eqref{equ:exp3_est2}. 
We present EXP3 in Algorithm~\ref{algo:EXP3}.
\begin{algorithm}[t]
	\centering
	\caption{Exponential-weight Algorithm for Exploration and Exploitation (Exp3)}
	\begin{algorithmic}[1]
		\STATE \textbf{Input: } $T, K, \eta$
		\STATE Set $\hat{S}_{0,a} = 0$ for all $a$.
		\FOR{$t = 1,\ldots, T$}
		\STATE Calculate $P_{ta}\leftarrow\frac{\exp\left(\eta\hat{S}_{t-1,a}\right)}{\sum_{a'\in\mathcal{A}}\exp\left(\eta \hat{S}_{t-1,a'}\right)}$ for all $a\in [K]$
		\STATE Sample $A_t$ from $P_t$ and receive reward $R_t$
		\STATE Calculate $\hat{S}_{t,a}\leftarrow\hat{S}_{t-1,a}+1-\frac{\mathbb{1}_{\{A_t = a\}}(1-R_t)}{P_{ta}}$
		\ENDFOR
	\end{algorithmic}
\label{algo:EXP3}
\end{algorithm}

Surprisingly, even though adversarial bandit problems look more difficult than stochastic bandit problems due to the strong power of the adversary, one can show that the adversarial regret for EXP3 algorithm has the same order as before, i.e. $O(\sqrt{KT\log T})$. 

\begin{thm}[Regret for EXP3 Algorithm]
\label{thm:EXP3}
	Let $r\in[0,1]^{T\times K}$. With learning rate $\eta = \sqrt{\log K/(KT)}$, we have
	\begin{align}
		\mathrm{Reg}_T(\mathrm{EXP3}, r) \leq 2\sqrt{KT\log K}.
	\end{align}
\end{thm}
\begin{proof}
	The proof for EXP3 is different than those for the stochastic bandit algorithms. We first define the expected regret relative to using action $a$ in $T$ rounds:
	\begin{align*}
		\mathrm{Reg}_{Ta} = \sum_{t = 1}^{T}r_{ta} - \mathbb{E}\left[\sum_{t=1}^{T}R_t\right].
	\end{align*}
	According to the definition of the adversarial bandit regret, the final result will follow if we can bound $R_{Ta}$ for every $a\in\mathcal{A}$. 
	It's not hard to show $\mathbb{E}[\hat{S}_{T,a}] = \sum_{t=1}^{T}r_{ta}$ and $\mathbb{E}_t[R_t] = \sum_{a\in\mathcal{A}}P_{ta}r_{ta} = \sum_{i = 1}^{K}P_{ta}\mathbb{E}[\hat{R}_{ta}]$ hold using the definition of $\hat{R}_{ta}$. 
	Then we can re-write $R_{Ta}$ as $\mathbb{E}\left[\hat{S}_{T,a}-\hat{S}_T\right]$, where $\hat{S}_T = \sum_{t=1}^{T}\sum_{a\in\mathcal{A}}P_{ta}\hat{R}_{ta}$. 
	Let $W_t = \sum_{a\in\mathcal{A}}\exp\left(\eta\hat{S}_{t,a}\right)$, $\hat{S}_{0,a} = 0$ and $W_0 = K$, 
	then one can show that 
	\begin{align*}
		\exp\left(\eta\hat{S}_{T,a}\right) \leq \sum_{a'\in\mathcal{A}}\exp\left(\eta\hat{S}_{T,a'}\right) = W_T = K\prod_{t=1}^{T}\frac{W_t}{W_{t-1}} = K\prod_{t=1}^{T}\sum_{a'\in\mathcal{A}}P_{ta'}\exp\left(\eta\hat{R}_{ta'}\right).
	\end{align*}
	We next bound the ratio term
	\begin{align*}
		\frac{W_t}{W_{t-1}} \leq 1+\eta\sum_{a'\in\mathcal{A}}P_{ta'}\hat{R}_{ta'}+\eta^2\sum_{a\in\mathcal{A}}P_{ta'}\hat{R}_{ta'}^2 \leq \exp\left(\eta\sum_{a'\in\mathcal{A}}P_{ta'}\hat{R}_{ta'}+\eta^2\sum_{a'\in\mathcal{A}}P_{ta'}\hat{R}_{ta'}^2\right),
	\end{align*}
	using inequalities $e^x\leq 1+x+x^2$ for $x\leq 1$ and $1+x\leq e^x$ for $x\in\mathbb{R}$.
	
	Combining with previous results, we have
	\begin{align*}
		\exp\left(\eta\hat{S}_{T,a}\right) \leq K\exp\left(\eta\hat{S}_T+\eta^2\sum_{t=1}^{T}\sum_{a'\in\mathcal{A}}P_{ta'}\hat{R}_{ta'}^2\right).
	\end{align*}
	Taking logarithm on both sides and rearranging give us
	\begin{align}
		\hat{S}_{T,a} - \hat{S}_T \leq \frac{\log K}{\eta}+\eta\sum_{t=1}^{T}\sum_{a'\in\mathcal{A}}P_{ta'}\hat{R}_{ta'}^2. \label{equ:EXP3_proof}
	\end{align}
	To bound $R_{Ta}$, we only need to bound the expectation of the second term in above. By standard (conditional) expectation calculations, one can get 
	\begin{align*}
		\mathbb{E}\left[\sum_{t=1}^{T}\sum_{a'\in\mathcal{A}}P_{ta'}\hat{R}_{ta'}^2\right] \leq TK.
	\end{align*}
	By substituting above inequality into \eqref{equ:EXP3_proof}, we get 
	\begin{align}
		\mathrm{Reg}_{Ta}\leq \frac{\log K}{\eta}+\eta TK = 2\sqrt{KT\log K},
	\end{align}
	where we choose $\eta = \sqrt{\log K/(TK)}$.
	By definition, the overall regret $\mathrm{Reg}_T(\mathrm{EXP3}, r)$ has the same upper bound as above. 
\end{proof}
We just proved the expected regret of EXP3. 
However, if we consider the distribution of the random regret, EXP3 is not good enough. 
Define the random regret as $\widehat{\text{Reg}}_T = \max_{a\in\mathcal{A}}\sum_{t=1}^{T}r_{ta} - \sum_{t=1}^{T}R_t$. 
One can show that for all large enough $T$ and reasonable choices of $\eta$, there exists a bandit such that the random regret of EXP3 satisfies $\mathbb{P}(\widehat{\text{Reg}}_T\geq T/4)\geq c>0$, where $c$ is a constant.
That means EXP3 sometimes can incur linear regret with non-trivial probability, which makes EXP3 unsuitable for practical problems. 
This phenomenon is caused by the high variance of the regret distribution. 
In next section, we will discuss how to resolve this problem by slightly modifying EXP3.

\paragraph{EXP3-IX (EXP3 with Implicit eXploration).}
We have learned that small $P_{ta}$ terms can cause enormous variance on the reward estimator, which then leads to high variance on the regret distribution. 
Thus, EXP3-IX~\citep{neu2015explore} redefines the loss-estimator as
\begin{align}
	\hat{Y}_{ta} = \frac{\mathbb{1}_{\{A_t = a\}}Y_t}{P_{ta}+\gamma},
\end{align}
where $Y_t = 1-R_t$ denotes the loss at round $t$ and $\gamma>0$.
$\hat{Y}_{ta}$ is a biased estimator for $y_{ta} = 1-r_{ta}$ due to $\gamma$, but the variance can be reduced.
An optimal choice for $\gamma$ needs to balance the bias and variance.
Other than this slight change on the loss estimator, the remaining procedures remain the same as EXP3. The name of 'IX' (Implicit eXploration) can be justified by the following argument:
\begin{align*}
	\mathbb{E}_t[\hat{Y}_{ta}] = \frac{P_{ta}y_{ta}}{P_{ta}+\gamma} \leq y_{ta}.
\end{align*}
The effect of adding a $\gamma$ to the denominator is that EXP3-IX tries to decrease the large losses for some actions, so that such actions can still be chosen occasionally.
As a result, EXP3-IX explores more than EXP3.  \citet{neu2015explore} has proved the following high probability regret bound for EXP3-IX.
\begin{thm}[Regret for EXP3-IX Algorithm]
	With $\eta = 2\gamma = \sqrt{\frac{2\log K}{KT}}$, \emph{EXP3-IX} guarantees that
	\begin{align}
		\widehat{\mathrm{Reg}}_T(\emph{EXP3-IX}) \leq 2\sqrt{2KT\log K} + \left(\sqrt{\frac{2KT}{\log K}}+1\right)\log(2/\delta)
	\end{align}
	with probability at least $1-\delta$.
\end{thm}

\subsubsection{Lower Bound for MAB Problems}
We have discussed two types of bandit models and their corresponding algorithms in regret minimization.
Then a natural question are: what is the minimal regret bound we can hope for?
To answer this question, we will introduce two types of lower bound results: minimax lower bound and instance dependent lower bound. 
Both of them are useful for describing the hardness of a class of bandit problems and are often used to evaluate the optimality of an existing algorithm.
For example, suppose the worst-case regret of a policy $\pi$ matches the minimax lower bound up to a universal constant, we say that the policy $\pi$ is minimax-optimal. 

\paragraph{Minimax Lower Bounds. }
We consider a Gaussian bandit environment, in which the reward for every arm is Gaussian-distributed. 
We denote the class of Gaussian bandits with unit variance by $\mathcal{E}_\mathcal{N}^K(1)$ and use $\mu\in\mathbb{R}^K$ as the reward mean vector. 
In particular, $\nu_\mu\in\mathcal{E}_\mathcal{N}^K(1)$ is a Gaussian bandit for which the $i$th arm has reward distribution $\mathcal{N}(\mu_i, 1)$. 
The following result provides a minimax lower bound for the Gaussian bandit class $\mathcal{E}_\mathcal{N}^K(1)$.

\begin{thm}[Minimax Lower Bound for Gaussian Bandit Class]
	\label{thm:minimax_mab}
	Let $K>1$ and $T\geq K-1$. For any policy $\pi$, there exists a mean vector $\mu\in[0,1]^K$ such that
	\begin{align}
		\operatorname{Reg}_T(\pi, \nu_\mu)  = \Omega(\sqrt{KT}).
	\end{align}
\end{thm}
\begin{proof}
	To prove the lower bound, we start with constructing two bandit instances that are very similar to each other and hard to  distinguish. 
	Let $\mu = (\Delta, 0, 0, \ldots, 0)$ denote the mean vector for the first unit variance Gaussian bandit. 
	We use $\mathbb{P}_\mu$ and $\mathbb{E}_\mu$ to denote the probability and expectation induced by environment $\nu_\mu$ and policy $\pi$ up to time $T$.
	To choose the second environment, let
	\begin{align*}
		i = \argmin_{j>1} \mathbb{E}_\mu[T_j(T)].
	\end{align*}
	Define the reward mean vector for the second bandit as $\mu'= (\Delta,0,\ldots,0,2\Delta,0,\ldots,0)$, where $\mu_i' = 2\Delta$.
	Decomposing the regret leads to
	\begin{align*}
		{\text{Reg}}_T(\pi, \nu_\mu) &\geq \mathbb{P}_\mu(T_1(T)\leq T/2)\frac{T\Delta}{2},\\
		{\text{Reg}}_T(\pi, \nu_{\mu'}) & > \mathbb{P}_{\mu'}(T_1(T) > T/2)\frac{T\Delta}{2}.
	\end{align*}
	Then, applying the Bretagnolle-Huber inequality, we get
	\begin{align*}
		\text{Reg}_T(\pi, \nu_\mu) + \text{Reg}_T(\pi, \nu_{\mu'}) \geq \frac{T\Delta}{4}\exp(-\text{KL}(\mathbb{P}_\mu, \mathbb{P}_{\mu'})).
	\end{align*}
	It remains to upper bound the KL-divergence term in above. 
	By divergence decomposition, one can show that 
	\begin{align*}
		\text{KL}(\mathbb{P}_\mu, \mathbb{P}_{\mu'}) = \sum_{i=1}^{K}\mathbb{E}_\mu[T_i(T)]\text{KL}(\mathbb{P}_i, \mathbb{P}_i') &= \mathbb{E}_\mu[T_i(T)]\text{KL}(\mathcal{N}(0,1), \mathcal{N}(2\Delta, 1))\\
		& = \mathbb{E}_\mu[T_i(T)]\frac{(2\Delta)^2}{2} \leq \frac{2T\Delta^2}{K-1}.
	\end{align*}
	In above , we use $\mathbb{P}_i$ and $\mathbb{P}_i'$ denote the reward distribution of the $i$th arm in $\nu_\mu$ and $\nu_{\mu'}$, respectively. 
	For the last inequality, since $\sum_{j=1}^{K}\mathbb{E}_\mu[T_j(T)] = T$, it holds that $\mathbb{E}_\mu[T_i(T)]\leq \frac{T}{K-1}$. 
	Combining with previous results, we know that 
	\begin{align*}
		\text{Reg}_T(\pi, \nu_\mu) + \text{Reg}_T(\pi, \nu_{\mu'}) \geq \frac{T\Delta}{4}\exp\left(-\frac{2T\Delta^2}{K-1}\right). 
	\end{align*}
	Choosing $\Delta = \sqrt{(K-1)/4T}\leq 1/2$, the result follows.
\end{proof}
An algorithm is called minimax-optimal if its worst-case regret matches with the minimax lower bound.
\paragraph{Instance Dependent Lower Bounds.}
An algorithm with nearly minimax-optimal regret is not always preferred, since it may fail to take advantage of environments that are not the worst case.
In practice, what is more desirable is to have algorithms that are near minimax-optimal, while their performance gets better on “easier” instances~\cite{lattimore2020bandit}.
This motivates the study of instance dependent regret. 
In this section, we present two types of lower bound for instance dependent regret: one is asymptotic, the other is finite-time. 

We first define consistent policy and present the asymptotic instance-dependent lower bound result.
\begin{defini}[Consistent Policy]
A policy $\pi$ is consistent if over bandit environment $\mathcal{E}$ if for all bandits $\nu\in\mathcal{E}$ and for all $p>0$ it holds that 
\begin{align*}
    \mathrm{Reg}_T(\pi, \nu) = O(T^p) \text{ as } n\rightarrow \infty.
\end{align*}
\end{defini}

\begin{thm}[Asymptotic Instance Dependent Lower Bound for Gaussian Bandits~\citep{lattimore2020bandit}]
For any policy $\pi$ consistent over $K$-armed unit-variance Gaussian environments $\mathcal{E}_\mathcal{N}^K(1)$ and any $\nu\in\mathcal{E}_\mathcal{N}^K(1)$, it holds that
\begin{align*}
    \liminf_{T\rightarrow\infty}\frac{\mathrm{Reg}_T(\pi, \nu)}{\log T} \geq \sum_{i:\Delta_i>0}\frac{2}{\Delta_i}.
\end{align*}
\label{thm:asymp_ins_lower}
\end{thm}
A policy is called asymptotically optimal if the equality in above theorem holds. 
Interestingly, building on the similar idea of Theorem~\ref{thm:asymp_ins_lower}, one can also develop a finite-time instance dependent lower bound result. 
\begin{thm}[Instance Dependent Lower Bound for Gaussian Bandits~\citep{lattimore2020bandit}]
	Let $\nu\in\mathcal{E}_\mathcal{N}^K(1)$ be a $K$-armed Gaussian bandit with mean vector $\mu\in\mathbb{R}^K$ and suboptimality gaps $\Delta\in[0,\infty)^K$. 
	Define a bandit environment:
	\begin{align*}
		\mathcal{E}(\nu) = \{\nu'\in\mathcal{E}_\mathcal{N}^K(1): \mu_i(\nu')\in[\mu_i,\mu_i+2\Delta_i]\}.
	\end{align*}
	Suppose $C>0$ and $p\in(0,1)$ are constants and $\pi$ is a policy such that $\mathrm{Reg}_T(\pi, \nu')\leq CT^p$ for all $T$ and $\nu'\in\mathcal{E}(\nu)$. 
	Then the regret for instance $\nu$ is lower bounded by
	\begin{align}
		\mathrm{Reg}_T(\pi, \nu) \geq \frac{2}{(1+\varepsilon)^2}\sum_{i:\Delta_i>0}\left(\frac{(1-p)\log T+\log\left(\frac{\varepsilon\Delta_i}{8C}\right)}{\Delta_i}\right)^+,
	\end{align}
	where $(x)^+:=\max\{x,0\}$.
\end{thm}
The proof for above two instance dependent lower bounds are too technical that we do not want to go beyond.
Interested readers are referred to Chapter 16 in \cite{lattimore2020bandit} for details.

\subsection{Contextual Bandit}\label{sec:contextual}
In many real-world applications, some side information is available to facilitate decision making. For instance, in healthcare applications, it is important to make decisions based on contextual information such as the person's demographic information, genetic information, life history, biomarkers, and environmental exposures. Ignoring such contextual information may lead to suboptimal decision making. To this end, the contextual bandits setting was introduced to model the side information that determines the reward distribution. The term ``contextual bandit'' is due to \cite{langford2007epoch} but similar settings have been considered earlier under the names ``bandit problems with covariates'' \citep{clayton1989covariate,sarkar1991one,yang2002randomized} and ``bandit problems with side information'' \citep{wang2005arbitrary,goldenshluger2011note}. Let $\mathcal{C}$ be a context space for the side information. The contextual bandits setting considered in this section
is the same as the MAB described in Section \ref{sec:MAB} except for an extra contextual
information $C_t \in \mathcal{C}$ at each step $t$, respectively. We extend our modeling of the environment from $\nu = (P_{a}: a \in \mathcal{A})$ to $\nu = (P_{a, c}: a \in \mathcal{A}, c \in \mathcal{C})$, i.e. the distribution of rewards depends on both action and context. The trajectory is extended to a sequence of triples, $(A_t, R_t, C_t)$, which represents the action, rewards and context at step $t$. An instantiation of the contextual bandit protocol in the healthcare setting is given in Algorithm~\ref{alg:cb}. Contextual bandits enable precision medicine by taking appropriate actions for particular patients rather than the same action for a variety of patients. Many papers, like \cite{rindtorff2019biologically} on precision oncology with \textit{in vitro} data and \cite{zhou2019tumor} on cancer treatment with genetic data, have effectively applied contextual bandits to precision medicine. Other applications can be found in \cite{shrestha2021bayesian}.

Throughout the section, we discuss the setting where $C_t$ is presented adversarially, rather than assuming $C_t$ is sampled stochastically from an underlying distribution. For the stochastic setting, \cite{goldenshluger2013linear} gives an algorithm with some parametric assumptions and \cite{yang2002randomized} with non-parametric algorithm. In fact, the stronger stochastic assumption does not improve the worst-case regret bound over the adversarial setting in linear contextual bandit that we will introduce later. The stochastic setting does help when the instance-dependent regret bound is considered. 

\begin{algorithm}
	\caption{Contextual Bandit Framework in Healthcare}
	\centering
	\begin{algorithmic}
		\STATE \textbf{Input:} Available treatment options ($\mathcal{A}$), treatment period length $T$.
		\FOR{$t=1,\ldots,T$}
		\STATE Agent receives the context information ($C_t$) for the current patient.
		\STATE A treatment decision ($A_t$) is made based on the history and the context information.
		\STATE Record the post treatment health outcome ($R_t$) for the patient.
		\ENDFOR
	\end{algorithmic}
	\label{alg:cb}
\end{algorithm}

In this section, we review the online learning literature on bandit problems with adversarial context. We will first see a naive one bandit per context approach, which does not make any assumption on the data generating process. We will also introduce the linear contextual bandit setting with linear assumption on the reward distribution. In the end, we introduce some variants of the standard contextual bandits that address sparsity or different types of outputs.

\subsubsection{One bandit per context}

The simplest solution for contextual bandits is possibly one bandit per context (OBPC), which solves the MAB for each context separately. Given any algorithm that solves MAB, OBPC maintains a trajectory for each context. On each step $t$, if $C_t = c$, the agent applies the algorithm on the trajectory corresponding to $c$ and selects an action. Combined with the EXP3 algorithms, which we call EXP3 per Context, OBPC achieves a regret bound of $\sqrt{K \log (K)|\mathcal{C}|T} $. OBPC does not make any further assumption on the data generating process $\nu$, so it only works when $\mathcal{C}$ is finite and small. As pointed out in \cite{lattimore2020bandit}, the awareness of the context information improves the performance over the non-contextual algorithm only when the context actually leads to a non-stationary environment and causes a large gap between the best context-aware policy
$\sum_{c \in \mathcal{C}} \max _{a \in[K]} \sum_{t \in[n]: X_{t}=c} R_{t,a}$ and non-context policy $\max _{a \in[K]} \sum_{t=1}^{n} R_{t,a}$.

\subsubsection{Bandits with expert advice}

OBPC fails when the context space is large and each context does not receive enough samples. In practice, context space has some internal structure that allows information sharing among different contexts. 
For instance, there can be thousands of categorical variables in precision health applications and many of them may be irrelevant to the reward distribution. Thus, grouping context based on these variables can reduce the context space while keeping the same performance.

Let $\mathcal{P} \subset 2^{\mathcal{C}}$ be a partition over context space, i.e. any $P, P' \in \mathcal{P}$, $P\cap P' = \emptyset$ and $\cup_{P \in \mathcal{P}}P = \mathcal{C}$. The above structural information can be seen as restricting the policy space within 
$$
    \Pi(\mathcal{P})=\left\{\pi: \mathcal{C} \rightarrow[K] ; \forall c, c^{\prime} \in \mathcal{C} \text { s.t. } c, c^{\prime} \in P \text { for some } P \in \mathcal{P}, \pi(c)=\pi\left(c^{\prime}\right)\right\}.
$$
In other words, we represent the structural information as a restricted policy space that is expected to be essentially smaller than the whole policy space. To allow more flexibility in structural information, we study the general restricted policy space $\Pi \subset \{\pi: \mathcal{C}\mapsto [K]\}$. The aim of agent is now to compete against a fixed class of policies, instead of competing against a set of actions which is a special case of the former. 

The recommendations of the policies in the restricted set can be thought of as ``expert advice''. We can therefore design an algorithm to use such advice by extending the EXP3 algorithm. The extended algorithm EXP4 (Exponential weighting for Exploration and Exploitation with Experts) is given in Algorithm \ref{algo:exp4}. Following the idea of EXP3, EXP4 first uses the same importance-weighted estimator for each action, i.e.
$
    \hat{R}_{t,i}=1 - {\mathbb{1}_{\{A_{t}=i\}} (1- R_{t})}/{P_{t i}}.
$
Then EXP4 estimates the reward for each expert (policy)
$$
    \tilde{R}_{t, \pi}=\sum_{i \in[K]} \mathbb{1}\left\{\pi\left(C_{t}\right)=i\right\} \hat{R}_{t,i}, 
$$
and keeps a distribution $Q_{t, \pi}$ over the policy space. The distribution is updated using exponentially weighting of the rewards:
$$
    Q_{t+1, \pi}=\exp \left(\eta R_{t, \pi}\right) Q_{t,\pi} / \sum_{\pi^{\prime} \in \Pi} \exp \left(\eta R_{t, \pi^{\prime}}\right) Q_{t,\pi}, 
$$
where $\eta \in (0, 1]$ is the learning rate.
\begin{algorithm}[t]
	\centering
	\caption{Exponential-weight Algorithm for Exploration and Exploitation with Experts (Exp4)}
	\begin{algorithmic}[1]
		\STATE \textbf{Input: } $T, K$, hyper-parameter $\eta \in (0, 1]$ and policy space $\Pi$
		\STATE Set the distribution $Q_{1, \pi} = 1/|\Pi|$ for all $\pi \in \Pi$
		\FOR{$t = 1,\ldots, T$}
		\STATE Receive context $C_t$
		\STATE For all $i \in [K]$, compute $P_{t,i} = \sum_{\pi \in \Pi} Q_{t, \pi} \mathbb{1}\{\pi(C_t) = i\}$
		\STATE Sample $A_t$ from $P_t$ and receive reward $R_t$
		\STATE Set $\hat R_{t,i} = 1 - \frac{\mathbb{1}\{A_t = i\}}{P_{t,i}}(1-R_t)$
		\STATE Propagate the rewards to the experts: $\tilde{R}_{t, \pi} = \sum_{i \in [K]} \mathbb{1}\{\pi(C_t) = i\} \hat R_{t,i}$ for all $\pi \in \Pi$
		\STATE Update $Q_{t+1, \pi} = {\exp(\eta \tilde{R}_{t, \pi})Q_{t,\pi}}/{\sum_{\pi' \in \Pi}\exp(\eta \tilde{R}_{t, \pi'})Q_{t,\pi}}$ for all $\pi \in \Pi$
		\ENDFOR
	\end{algorithmic}
	\label{algo:exp4}
\end{algorithm}

We define regret not with respect to the best action but with respect to the best expert:
$$
    \text{Reg}_{T}=\mathbb{E}\left[\max_{\pi \in \Pi} \sum_{t=1}^{T} \bm{r}_t^T\pi(C_t)-\sum_{t=1}^{n} R_{t}\right],
$$
where $\bm{r}_t = (r_{t,i})_i$ is the $K$-dimensional vector of rewards of the individual actions and $\pi(C_t)$ is the $K$-dimensional vector of the probability of selecting each action using policy $\pi$ under context $C_t$.
\begin{thm}
Let $\eta = \sqrt{2\log(|\Pi|/(TK))}$. The regret of \emph{EXP4} satisfies 
$$
    \mathrm{Reg}_T(\mathrm{EXP4}) \leq \sqrt{2\log(|\Pi|)KT}.
$$
\end{thm}

\begin{proof}
	Similarly to the way we show (\ref{equ:EXP3_proof}) in the proof in Theorem \ref{thm:EXP3}, we can also show that the for any $\pi^* \in \Pi$:
	\begin{equation}
	    \sum_{t=1}^{T} \tilde{R}_{t, \pi^{*}}-\sum_{t=1}^{T} \sum_{\pi \in \Pi} Q_{t, \pi} \tilde{R}_{t, \pi} \leq \frac{\log (|\Pi|)}{\eta}+\frac{\eta}{2} \sum_{t=1}^{T} \sum_{\pi \in \Pi} Q_{t, \pi}\left(1-\tilde{R}_{t, \pi}\right)^{2}. \label{equ:exp4_proof}
	\end{equation}
	Let $\hat{\bm{R}}_t = (\hat{R}_{t, i})_{i = 1}^{K}$ and $\tilde{\bm{R}}_t = (\tilde{R}_{t, \pi})_{\pi \in \Pi}$ denote the vector of $R_{t, i}$'s and $\tilde{R}_{t, \pi}$'s. Since the estimator is unbiased, $\mathbb{E}_t[\hat R_t] = \bm{r}_t$ and letting $\bm{E}^{(t)}$ be the $|\Pi| \times K$ matrix with each row corresponding to a $\pi \in \Pi$ that gives the row vector $(\mathbb{1}\{\pi(C_t) = i\})_{i = 1}^K$, we have
	$$
	    \mathbb{E}_{t}[\tilde{\bm{R}}_{t}]=\mathbb{E}_{t}[\bm{E}^{(t)} \hat{\bm{R}}_{t}]=\bm{E}^{(t)} \mathbb{E}_{t}[\hat{\bm{R}}_{t}]=\bm{E}^{(t)} \bm{r}_t.
	$$
	Taking expectation over both side of Eq. (\ref{equ:exp4_proof}) and using the fact that $Q_{t, \pi}$ is $\mathcal{F}_t$-measurable leads to 
	\begin{equation}
	    \text{Reg}_{T} \leq \frac{\log (|\Pi|)}{\eta}+\frac{\eta}{2} \sum_{t=1}^{T} \sum_{\pi \in \Pi} \mathbb{E} Q_{t, \pi}\left(1-\hat{R}_{t, \pi}\right)^{2}. \label{equ:exp4_proof2}
	\end{equation}
	Let $\hat Y_{t,i} = 1-\hat R_{t,i}$, $y_{t,i} = 1 - r_{t,i}$ and $\tilde{Y}_{t,\pi} = 1-\tilde{R}_{t,\pi}$ for all $i \in [K]$ and $\pi \in \Pi$. Define vectors $\tilde{\bm{Y}}_t$ and $\hat{\bm{Y}}_t$ similarly.  Note that $\tilde{\bm{Y}}_t = \bm{E}^{(t)} \hat{\bm{Y}}_t$ and let $A_{t,i} = \mathbb{1}\{{A}_t = i\}$, which gives us $\hat Y_{t,i} = A_{t,i}y_{t,i} / P_{t,i}$ and 
	$$
	    \mathbb{E}_{t}\left[\tilde{Y}_{t, \pi}^{2}\right]=\mathbb{E}_{t}\left[\left(\frac{E_{\pi, A_{t}}^{(t)} y_{t, A_{t}}}{P_{t, A_{t}}}\right)^{2}\right]=\sum_{i=1}^{K} \frac{\left(E_{\pi, i}^{(t)} y_{t, i}\right)^{2}}{P_{t, i}} \leq \sum_{i=1}^{K} \frac{E_{\pi, i}^{(t)}}{P_{t, i}}.
	$$
	Thus, by the definition of $P_{t,i}$,
	$$
	    \mathbb{E}[\sum_{\pi \in \Pi} Q_{t,\pi}(1-\tilde{R}_{t,\pi})] \leq \left[\sum_{\pi \in \Pi} Q_{t, \pi} \sum_{i=1}^{K} \frac{E_{\pi, i}^{(t)}}{P_{t, i}}\right] = K.
	$$
	Substituting into Eq. (\ref{equ:exp4_proof2}) gives us
	$$
	    \text{Reg}_T \leq \frac{\log(|\Pi|)}{\eta} + \frac{\eta T K}{2} \leq \sqrt{2T K \log(|\Pi|)},
	$$
	when $\eta = \sqrt{2\log(|\Pi|) / (TK)}$.
\end{proof}

Now we take a look at the context space partition example introduced before. Assume we have a $M$-partition of the context space $\mathcal{C}$. The regret bound is reduced from $\sqrt{2|\mathcal{C}|\log(K)KT}$ to $\sqrt{2M\log(K)KT}$ with the structural information.

\subsubsection{Linear bandits and contextual linear bandits}

One bandit per context method learns a distinct environment for each context, which fails when the context space is large or even infinite. Many healthcare applications, especially in mobile health, have continuous context including heart rate, sleeping time, etc. To model such applications with efficient algorithms, we need to assume some nice relationship between the context and environment. The simplest and most frequently used model of this kind is the contextual linear bandit.

Before we introduce the setting for contextual linear bandit, we first define linear bandit. In linear bandit, we do not directly model contexts but assume that the action space $\mathcal{A}_t \subset \mathbb{R}^d$ (we will soon reveal the reason behind choosing a time varying action space) for some positive integer $d$, the reward on step $t$ is sampled from a linear model:
\begin{equation}
    R_t = \left\langle\theta^*, A_{t}\right\rangle + \eta_t, \text{ for some } \theta^* \in \mathbb{R}^d, \label{equ:linear_bandit}
\end{equation}
and the noise $\eta_t$ is usually assumed to be subgaussian. Note that, depending on the choice of $\mathcal{A}_t$, linear bandits can subsume other bandit problems. For example, to capture the standard stochastic bandit environment with $d$ actions, we set $\mathcal{A}_t = \{e_1, \dots, e_d\}$ for unit vectors $(e_i)_i$. Note that this choice of action set is not time varying. The flexibility to choose time varying action sets is critical to capture contextual bandits. We simply let $\mathcal{A}_t = \{\psi\left(C_{t}, i\right) \in \mathbb{R}^d: i \in[K]\}$, for some known function $\psi: \mathcal{C} \times [K] \mapsto \mathbb{R}^d$. A common setup \citep{li2010contextual} considers one context per arm, where $\mathcal{C} \subset \mathbb{R}^{d \times K}$ and the function $\psi(c, i)$ extracts the $i$-th column of the matrix $c$. Another formulation assumes the same context $c_t \in \mathbb{R}^d$ for every arm but assumes each arm $i$ has a parameter $\theta_{i}^* \in \mathbb{R}^d$:
$
    R_t = \left\langle \theta_{i}^*, c_t \right\rangle + \eta_t,
$ when arm $i$ is chosen.
Note that this is a special case of the general formulation in (\ref{equ:linear_bandit}).
We can write $\theta^* = (\theta_1^{*T}, \dots, \theta_K^{*T})^T$ and $\mathcal{A}_t = \{(\mathbf{0}_{d(i-1)},C_{t}^T,\mathbf{0}_{d(K-i)})\}_{i = 1}^K \subset \mathbb{R}^{dK}$, where $\mathbf{0}_{n}$ is the all-zero row vector of dimension $n$. 

\paragraph{Linear Upper Confidence Bound (LinUCB).} 
The regret for linear bandit is defined as 
$$
    \text{Reg}_T = \mathbb{E}[\sum_{t = 1}^T \max_{a_t \in \mathcal{A}_t} \langle \theta^*, a_t \rangle - \sum_{t = 1}^T R_t].
$$

We have seen UCB for the regular bandit environment as a powerful algorithm to balance the exploration and exploitation. In this section, we introduce a new UCB algorithm called LinUCB for linear bandit setting \citep{auer2002using,dani2008stochastic,rusmevichientong2010linearly}. 

As the case in UCB, we follow the principle \textit{optimism in the face of uncertainty (OFU)}. Instead of directly building confidence set over the mean function $\mu_{a}$, we build confidence set $\Theta_t \subset \mathbb{R}^d$ over the linear coefficient $\theta^*$ that satisfies the following two properties: 1) $\Theta_t$ contains the optimal $\theta^*$ with a high probability; 2) the set $\Theta_t$ should be as small as possible. 

For that purpose, we consider the regularised least-square estimator up to step $t$,
\begin{equation}
\label{equ:ls_est}
\hat{\theta}_{t}=\operatorname{argmin}_{\theta \in \mathbb{R}^{d}}\left(\sum_{s=1}^{t}\left(R_{s}-\left\langle\theta, A_{s}\right\rangle\right)^{2}+\lambda\|\theta\|_{2}^{2}\right),
\end{equation}
where $\lambda > 0$ is the penalty factor. Equ. (\ref{equ:ls_est}) gives 
\begin{equation}
\label{equ:vt}
    \hat{\theta}_{t}=V_{t}^{-1} \sum_{s=1}^{t} A_{s} R_{s},\quad  \text{ where }
    V_0 = \lambda I_d \text{ and } V_{t}=V_{0}+\sum_{s=1}^{t} A_{s} A_{s}^{\top}.
\end{equation}

Then the confidence set centering at $\hat \theta_t$ is defined by 
\begin{equation}
    \Theta_{t+1} = \{\theta \in \mathbb{R}^d : \|\theta - \hat \theta_{t}\|_{V_{t}}^2 \leq \beta_{t+1}\}, \label{equ:linucb_cf}
\end{equation}
where $(\beta_t)_t$ is an increasing sequence of constants that controls the level of confidence with $\beta_1 \geq 1$ as $(V_t)_t$ also have increasing eigenvalues. Lemma \ref{lemma:linucb_cf} shows the correctness of the confidence set defined in (\ref{equ:linucb_cf}). LinUCB chooses actions that maximize the expected reward over all the possible $\theta$'s in the confidence set. Details are given in Algorithm \ref{algo:LinUCB}.

\begin{algorithm}[t]
	\centering
	\caption{Linear Upper Confidence Bound (LinUCB)}
	\begin{algorithmic}[1]
		\STATE \textbf{Input: } $\delta$, $d$, $(\mathcal{A}_t)_t \subset \mathbb{R}^d$, $(\beta_t)_t$, $\lambda$. 
		\FOR{$t = 1,\ldots, T$}
		\STATE Calculate $\hat \theta_t$ using (\ref{equ:ls_est}).
		\STATE Compute the confidence set $\Theta_t = \{\theta \in \mathbb{R}^{d}:\|\theta-\hat{\theta}_{t-1}\|_{V_{t-1}}^{2} \leq \beta_{t}\}$.
		\STATE Play $A_t = \argmax_{a \in \mathcal{A}_t} \max_{\theta \in \Theta_t} \langle\theta, a\rangle$.
		\STATE Receive reward $R_t$ and update $V_t$ using (\ref{equ:vt}).
		\ENDFOR
	\end{algorithmic}
\label{algo:LinUCB}
\end{algorithm}

\begin{lemma}
\label{lemma:linucb_cf}
Let $\delta \in (0, 1)$. Then, with a probability at least $1-\delta$, it holds that for all $t \in \mathbb{N}$, $\theta^* \in \Theta_t$ defined in (\ref{equ:linucb_cf}) with any sequence $(\beta_t)_t$ satisfying
$$
   \beta_t \leq  \sqrt{\lambda}\left\|\theta^{*}\right\|_{2}+\sqrt{2 \log \left(\frac{1}{\delta}\right)+d\log \left(\frac{\operatorname{det} V_{t}}{\lambda^{d}}\right)}.
$$
\end{lemma}

Now using the confidence set in Lemma \ref{lemma:linucb_cf}, we can show the following regret bound under some mild assumption on the boundedness of the action space. We refer reader to Part V of \cite{lattimore2020bandit} for the proof of Lemma \ref{lemma:linucb_cf}.
\begin{thm}
\label{thm:LinUCB}
Assuming $\max _{t \in[n]} \sup _{a, b \in \mathcal{A}_{t}}\left\langle\theta^{*}, a-b\right\rangle \leq 1$ and $\|a\|_{2} \leq L \text { for all } a \in \bigcup_{t=1}^{n} \mathcal{A}_{t}$, using the confidence set defined in Lemma \ref{lemma:linucb_cf}, we have with a probability at least $1-\delta$, the expected regret of LinUCB is bounded by
$$
    \operatorname{Reg}_T(LinUCB, \nu) \leq Cd\sqrt{T}\log(L/\delta),  \text{ for some } C > 0.
$$
\end{thm}
\begin{proof}
We first analyze the regret on each step. Let $\title{\theta}_t \in \Theta_t$ be the parameter that maximize the expected reward. Let $A_t^*$ be the optimal action given context $C_t$. Then, using the fact that $\theta^* \in \Theta_t$, we have 
$
    \langle \theta^*, A_t^* \rangle \leq \langle \title{\theta}, A_t \rangle
$.
Using Cauchy-Schwarz inequality, we have 
$$
    r_t = \langle\theta_{*}, A_{t}^{*}-A_{t}\rangle \leq \langle\title{\theta}_t - \theta_{*}, A_{t}\rangle \leq \|A_{t}\|_{V_{t-1}^{-1}}\|\tilde{\theta}_{t}-\theta_{*}\|_{V_{t-1}} \leq 2 \|A_t\|_{V_{t-1}^{-1}}\sqrt{\beta_t}.
$$

Then the total regret is given by 
\begin{equation}
\text{Reg}_T \leq \sum_{t = 1}^T 2\left\|A_{t}\right\|_{V_{t-1}^{-1}} \sqrt{\beta_{t}} \leq \sqrt{T \sum_{t = 1}^T r_t^2} \leq \sqrt{T  \beta_T\sum_{t=1}\|A_t\|^2_{V_{t-1}^{-1}}}.
\label{equ:regret_LinUCB}
\end{equation}
Using the elliptical potential lemma \citep{carpentier2020elliptical},
$
    \sum_{t=1}^{T}\left(1 \wedge\left\|A_{t}\right\|_{V_{t-1}^{-1}}^{2}\right) \leq 2 \log \left({\operatorname{det} V_{T}}/{\operatorname{det} V_{0}}\right).
$
The results follow by plugging this into (\ref{equ:regret_LinUCB}) with the chosen $\beta_T$ and the fact that 
\begin{equation*}
    \operatorname{det}\left(V_{T}\right)=\prod_{i=1}^{d} \lambda_{i} \leq\left(\frac{1}{d} \operatorname{trace} V_{T}\right)^{d} \leq\left(\frac{\operatorname{trace} V_{0}+T L^{2}}{d}\right)^{d},
\end{equation*}
where $\lambda_1, \dots, \lambda_d$ are eigenvalues of $V_t$. 

\end{proof}

LinUCB gives a regret bound that scales with $d$. However, this is not the optimal rate when the action set is finite. 
\cite{auer2002using} 
proposed SupLinUCB which maintains a master algorithm that achieves a regret bound of $\sqrt{dT\log(TK)}$ when the action space is finite.

\paragraph{LinUCB for contextual bandits.} \cite{chu2011contextual} applied the LinUCB to the contextual bandit problems. To be specific, they assume a context and a unknown parameter $\theta$ for each arm $a$ and the reward is generated by
$$
    R_t = C_{t, A_t}^{\top} \theta_{A_t}^* + \eta_t, \text{ where } C_{t, a}, \theta_a^* \in \mathbb{R}^d.
$$
 Applying Theorem \ref{thm:LinUCB} gives us the regret bound of $Kd\sqrt{T}$. \cite{li2010contextual} showed a good performance of the contextual bandit with LinUCB on News Article Recommendation task.

\subsubsection{Sparse LinUCB}
So far we have achieved regret bound of $d\sqrt{T\log(T)}$, which scales linearly with $d$. However, the bound can be vacuous for high-dimensional context ($d \gg T$). Such high dimensional problems can occur in healthcare applications. For example,  \cite{bastani2020online} incorporates the high-dimensional genetic profile and medical records with contextual bandits to design patient's optimal medication dosage. Following the idea of UCB, the key is to construct a confidence set that accounts for the sparsity. To tackle the problem, we introduce a powerful tool that converts online prediction problem to confidence set estimation, with which a UCB-style algorithm can be designed using any online predicting method and we can analyze its regret in a unified way. 

\paragraph{Online to Confidence Set Conversion.} There is a conversion from the online prediction to confidence sets. 
We consider online linear regression problem with a squared loss, which has been well-explored in the past decade \citep{huang2008adaptive,javanmard2014confidence}. The agent interacts with the environment in the following manner where in each round $t$:
\begin{enumerate}
    \item The environment chooses a context $C_t \in \mathbb{R}$ and $A_t \in \mathbb{R}^d$ in an arbitrary way. 
    \item The value of $A_t$ is revealed to the agent.
    \item The agent generate a prediction $\hat R_t$.
    \item The environment reveals $R_t$ to the agent as well as the loss $(R_t - \hat R_t)^2$.
\end{enumerate}


The regret of an agent with respective to the best linear predictor is given by
$$
    \rho_T = \sum_{t = 1}^T (R_t -\hat R_t)^2 - \inf_{\theta \subset \mathbb{R}^d}\sum_{t = 1}^T(R_t - \langle\theta, A_t\rangle)^2.
$$

\begin{thm}[Online-to-Confidence-Set Conversion \citep{abbasi2012online}]
\label{thm:OL2Confi_Set}
Let $\delta \in (0, 1)$ and assume that $\theta^* \in {\Theta}$, noise $\eta_t$'s are $R$-sub-Gaussian and $\rho_t \leq B_t$. Let 
$$ 
\beta_{t}(\delta)=1+2 B_{t}+32 R^2 \log \left(\frac{R\sqrt{8}+\sqrt{1+B_{t}}}{\delta}\right).
$$
Then
\begin{equation}
    \Theta_{t+1} = \left\{ \theta \in \mathbb{R}^d: \sum_{s = 1}^t(\hat R_s - \langle\theta, A_s\rangle)^2 \leq \beta_t(\delta) \right\}, \label{equ:online_confi_set}
\end{equation}
satisfies $\mathbb{P}(\theta^* \in \Theta_{t+1} \forall t \in \mathbb{N}) \geq 1-\delta$.
\end{thm}

\begin{algorithm}[t]
	\centering
	\caption{Online linear predictor UCB}
	\begin{algorithmic}[1]
		\STATE \textbf{Input: } $\delta \in (0, 1)$, an online predictor with a regret bound $B_t$
		\FOR{$t = 1,\ldots, T$}
		\STATE Receive action set $\mathcal{A}_t$
		\STATE Compute the confidence set $\Theta_t$ using (\ref{equ:online_confi_set})
		\STATE Play $A_t = \argmax_{a \in \mathcal{A}_t} \max_{\theta \in \Theta_t} \langle\theta, a\rangle$ and receive reward $R_t$
		\STATE Feed $A_t$ to the online linear predictor and obtain prediction $\hat R_t$
		\STATE Feed $R_t$ and loss to the linear predictor as feedback.
		\ENDFOR
	\end{algorithmic}
\label{algo:OLR-UCB}
\end{algorithm}

Theorem \ref{thm:OL2Confi_Set} provides a tool for constructing confidence set for the prediction of any online learning algorithm with a regret guarantee. One can develop the Online linear predictor UCB (OLR-UCB) as shown in Algorithm \ref{algo:OLR-UCB}. 

Let $\|\theta^*\|_2 \leq m_2$. Using Theorem \ref{thm:OL2Confi_Set} and the standard regret decomposition for UCB-style algorithm, one can show the following regret bound:
$$
    {R}_{T} \leq \sqrt{8 d T\left(m_{2}^{2}+\beta_{T-1}(\delta)\right) \log \left(1+\frac{T}{d}\right)}.
$$

\paragraph{Application to sparse online linear prediction.} Note that our goal is to derive a regret bound with a lower dependency on the ambient dimension $d$. In the standard regression problem, learnability of high-dimensional data relies on the sparsity assumption which assumes the number of non-zero dimensions is less or equal to a positive integer $s$.
\begin{assumption}(sparse parameter)
The true parameter $\theta^*$ satisfies $\|\theta^*\|_0 \leq s$ for some $s > 0$.
\end{assumption}

To this end, we adopt SeqSEW \citep{gerchinovitz2011sparsity} as the online linear prediction algorithm, which achieves the online learning regret $\rho_T = \mathcal{O}( s\log(T))$. Combined with Theorem~\ref{thm:OL2Confi_Set}, we have the following result:
\begin{thm}
With a high probability, the we bound the regret of OLR-UCB with SeqSEW as the online predictor by
\begin{equation}
    \mathrm{Reg}_T = \tilde{\mathcal{O}}({\sqrt{dsT}}). \label{equ:sparse_lin}
\end{equation}
\end{thm}
This improves over the $d\sqrt{T}$ by LinUCB in the previous section when $s \ll d$. As pointed out by \cite{lattimore2020bandit}, for any algorithm, there exists an infinite action space such that $\mathrm{Reg}_T = \Omega(\sqrt{dsT})$. Thus, the bound in (\ref{equ:sparse_lin}) is minimax optimal.
Despite its optimality, our result still have the unavoidable dependence of $\sqrt{d}$. \cite{bastani2020online} derived a regret bound of $\mathcal{O}(K(s\log(d)\log(T))^2)$, when the action set is finite. More variants of high-dimensional sparse linear bandits are discussed in \cite{hao2020high}. 

\subsection{Offline Learning}\label{sec:offline}

So far we have been discussing bandit problems in the \emph{online learning} setting, where we are free to choose actions while collecting the dataset. In many real applications, collecting data in the online manner can be expensive and can cost a lot of time. In such situations we might want to evaluate a target policy using a large but fixed dataset that has been collected for years by a different policy. 
This setting is called \emph{offline learning}. 

Given a fixed contextual bandit environment, the value of a policy is defined as $v_{\pi} = \mathbb{E}_{a_t \sim \pi(c_t)} r_t$ 
Formally, offline bandits, also referred as \textit{off-policy evaluation} aims at evaluating the value $v_{\pi}$ 
of a new policy using a sequence of interactions $S_T = (c_t, a_t, r_t)_{t = 1}^T$, where $c_t$, $a_t$ and $r_t$ are the context, action and reward respectively at step $t$. The actions are chosen by the data-generating policy $\pi_{D}$, which is also referred as \emph{behavior policy} or \emph{logging policy}. In this section, 

One solution is to learn the reward and context distribution using a parametric or non-parametric method and then evaluate the new policy with the simulator by sampling context and rewards from the learnt distribution. 
Although this approach is straightforward, the modeling step is often very expensive and difficult, and more importantly, it often introduces modeling bias to the simulator, making it hard to justify reliability of the obtained evaluation results.

\cite{li2011unbiased} proposed a simple algorithm (Algorithm \ref{algo:Off-Unbias}) that uses an unbiased estimator of the policy value when the behavior policy picks arms uniformly at random. The method is to simply use all the interactions in the dataset that coincide with the action that the policy being evaluated would have taken. They show that $\hat{v}_{\pi}$ is an unbiased estimator of $v_{\pi}$ and with a high probability 
$
    ({\hat{v}_{\pi}} - v_{\pi})^2 = \tilde{\mathcal{O}}\left({{Kv_{\pi}}/{n}}\right),
$
where $v_{\pi}$ is the true expected value of the policy as defined above. They also successfully applied the off-policy evaluation to the News Article Recommendation system. 
\cite{mary2014improving} improved upon Algorithm~\ref{algo:Off-Unbias} via bootstrapping techniques. They give a bootstrapped estimator for any quantile of the distribution of the estimated value $\hat v_{\pi}$ that allows some evaluation of the uncertainty.

\begin{algorithm}[ht]
	\centering
	\caption{Policy Evaluator}
	\begin{algorithmic}[1]
		\STATE \textbf{Input: } a policy $\pi$; stream of interactions $S_T$ of length $T$
		\STATE $h_0 \leftarrow \emptyset$, $\hat V_{\pi} \leftarrow 0$ and $N \leftarrow 0$
		\FOR{$t = 1,\ldots, T$}
		\STATE Get the $i$-th interaction $(c_i, a_i, r_t)$ from $S_T$
		\IF{$\pi(h_{t-1}, c_t) = a_t$}
		\STATE $h_t \leftarrow h_{t-1} \cup (c_t, a_t, r_t)$
		\STATE $\hat V_{\pi} \leftarrow \hat V_{\pi} + r_t$ and $N \leftarrow N + 1$
		\ENDIF
		\ENDFOR
		\STATE Output: $\hat{v}_{\pi} = \hat V_{\pi} / N$
	\end{algorithmic}
\label{algo:Off-Unbias}
\end{algorithm}


\paragraph{Minimax rate lower bound.} To understand how good our algorithms perform, we start with establishing a minimax lower bound that characterizes the inherent hardness of the off-policy value estimation problem. Let $R_{\max}, \sigma\in \mathbb{R}^+$. 
Define the class of reward distributions $\mathcal{V}(\sigma, R_{\operatorname{max}})$ with bounded mean and variance as 
$$
    \mathcal{V}(\sigma, R_{\operatorname{max}}) \coloneqq \{\nu = (P_{a, c})_{a,c\in\mathcal{A}\times \mathcal{C}}: 0 \leq \mathbb{E}_{R \sim P_{a, c}}[R] \leq R_{\max} \text{ and } \operatorname{Var}[R \mid a, c] \leq \sigma^2, \ \forall a,c\in\mathcal{A}\times \mathcal{C}\}.
$$
Any estimator $\hat v$ is a function that maps $(\pi, \pi_D, S_T)$ to an estimate of $v_{\pi}$. Let the context $C \sim \lambda$ with a density relative to Lebesgue measure. The minimax rate is defined as 
$$
    M_n(\pi, \lambda, \pi_D, \sigma, R_{\max}) \coloneqq \inf_{\hat v} \sup_{\nu \in \mathcal{V}(\sigma, R_{\operatorname{max}})} \mathbb{E}[(\hat v(\pi, \pi_D, S_T) - v_{\pi})^2],
$$
where the expectation is taken over the distribution of the dataset $S_T$ that depends on $(\pi, \lambda, \pi_D)$. 

A critical value that determines the hardness of the problem is 
$$
    V_1 \coloneqq \mathbb{E}_{C\sim \lambda, A\sim \pi_D(\cdot, C)}\left[{\pi^2(A, C)}/{\pi_D^2(A, C)} \right].
$$ 
\cite{wang2017optimal} (Corollary 1) gives the following lower bound for $R_n$:
\begin{equation}
    M_n(\pi, \lambda, \pi_D, \sigma, R_{\max}) \geq \frac{V_1(\sigma^2 + R_{\max}^2)}{700T}. \label{eq:minimax_offline}
\end{equation}

\paragraph{Importance sampling estimator.} \cite{wang2017optimal} give some analysis on the minimax rate of the off-policy evaluation given a known data-generating policy, denoted by $\pi_D$. We slightly abuse the notation and let $\pi(a, c)$ denote the probability of choose $a$ using policy $\pi$ under the context $c$. They consider Importance Sampling (IS) estimator \citep{charles2013counterfactual} and Weighted Importance Sampling (WIS) estimator given by
$$
    \hat{v}_{\mathrm{IS}, \pi} = \frac{1}{T}\sum_{t = 1}^T \frac{\pi(a_t, c_t)}{\pi_D(a_t, c_t)}r_t \quad\text{ and } \quad
    \hat{v}_{\mathrm{WIS}, \pi}=\sum_{t=1}^{T} \frac{\frac{\pi( a_t, c_t)}{\pi_{D}( a_t, c_t)}}{\sum_{t'=1}^{T} \frac{\pi(a_{t'}, c_t)}{\pi_{D}(a_{t'}, c_t)}} r_{t} \text{, respectively.}
$$

It is shown in \cite{dudik2014doubly} that the IS estimator is unbiased and it achieves the minimax rate in (\ref{eq:minimax_offline}) when $n$ is sufficiently large. The WIS estimator is biased with a lower variance. It also enjoys the minimax optimality up to a logarithmic constant. 


\paragraph{Regression estimator.}
Another common method, so called regression estimator or plug-in estimator, simply learns the reward distribution and plugs it in. 
Let $\hat r:\mathcal{A} \times \mathcal{C} \mapsto \mathbb{R}$ is an estimator for the mean reward. The regression estimator is given by 
$$
    \hat{v}_{\mathrm{Reg}, \pi}:= \frac{1}{T} \sum_{t = 1}^T\sum_{a} \pi(a, c_t) \hat{r}(a, c_t).
$$
When the context space and action space are finite. One can use the simple sample average estimator $\hat r(a, c) = (\sum_{t = 1}^T \mathbb{1}(a_t = a, c_t = c) r_i) / (\sum_{t = 1}^T \mathbb{1}(a_t = a, c_t = c))$.
Interestingly, one can write the regression estimator using sample average as 
$$
    \hat{v}_{\operatorname{Reg}}=\frac{1}{T} \sum_{t=1}^{T} \frac{\pi\left(a_{t}, c_t\right)}{\hat{\pi}_{D}\left(a_{t}, c_t\right)} r_{t}, \text{ where } \hat{\pi}_{D}\left(a, c\right) = \frac{\sum_t \mathbb{1}(a_t = a, c_t = c)}{T}.
$$
There is a strong connection between IS and regression estimator. That is regression estimator simply replaces the $\pi_D$ with its empirical estimator. 
Regression estimator is biased while its variance is normally lower than IS \citep{dudik2011doubly, li2015toward}. Regression estimator is also shown to be minimax optimal when $n$ is sufficiently large. The simple plugin method using sample average is not applicable when context space is infinite. When more information on how rewards depend on contexts and actions, we can posit a parametric or non-parametric model of $\mathbb{E}[R \mid C, A]$ and fit it to obtain an estimator. Due to the difficulties of having a good estimation $\hat x$, which may suffer model-misspecification or high variance, pure regression estimator only works well in problem with finite actions. For more details, we refer reader to  a more recent paper \citep{kim2021doubly}. 

\paragraph{Doubly robust estimator.} Based on the above discussions about IS and Regression Estimator, it is natural to combine the both. Doubly Robust (DR) \citep{bang2005doubly,dudik2011doubly,jiang2016doubly} which is a combination of regression and IS estimator, and can achieve the low variance of regression and no (or low) bias of IS. The method is named by Doubly Robust because it is accurate if at least one of the estimator is accurate. DR estimator is defined as 
$$
    \hat{v}_{DR} = \frac{1}{T} \sum_{t} \left[\frac{\pi(a_t, c_t)}{\pi_D(a_t, c_t)}(r_t - \hat{r}(a_t, c_t)) + \sum_{a}\pi(a, c_t)\hat{r}(a, c_t)\right].
$$
Informally, the estimator uses $\hat{r}(a, c)$ as a baseline and corrects the baseline when the action more likely to be sampled from the target policy. DR enjoys both the low variance of Regression estimator and the low bias of the IS estimator and also matches the minimax lower bound in \ref{eq:minimax_offline}. \cite{wang2017optimal} further introduces the SWITCH estimator that takes account into the scale of the importance weight. When the importance weight is large, it switches to the regression estimator, which significantly reduces the variance while introducing only little bias. It performs better in the numerical experiments. 

\paragraph{Non-asymptotic regime.} All of the above estimators achieve minimax optimality when $n$ is sufficiently large. \cite{ma2021minimax} discussed regime when the sample size is not large. They showed that there is a fundamental statistical gap between algorithm with and without the knowledge of the behavior policy in this regime. They proposed a \textit{competitive ratio} that measures the MSE of an algorithm with unknown behavior policy relative to the lower bound of all algorithms with known behavior policy. Competitive ratio is always greater than 1. A competitive ratio closer to 1 indicates that the algorithm can performance as good as the algorithm with known behavior policy.
\cite{ma2021minimax} showed that regression estimator has minimax optimal competitive ratio of the rate is $\mathcal{O}(K)$, where $K$ is the number of actions. That means an algorithm without the knowledge of the behavior policy has to pay multiplicative factor of $K$ compared to that with the knowledge.

\section{Advanced Topics}\label{sec:advs}

We now review a selection of advanced topics in bandit algorithms that are relevant to applications in healthcare.

\subsection{Non-stationarity} \label{sec:non_stationary}
We have discussed two basic and extreme cases of bandit theory: stochastic bandit models and adversarial bandit models. In the former, the reward distribution for every arm is static whereas in the latter, it can change arbitrarily over time. 
In this section, we describe a more practical setting that sits in the ``middle'' of above two extreme models: non-stationary bandits.
Specifically, the reward distributions change over time, but the amount of changes cannot be as arbitrary as the adversarial setting. 

For a non-stationary multi-armed bandit environment $\nu$, we define the reward mean for arm $a$ at time $t$ by $\mu_{a,t}(\nu)$ or $\mu_{a,t}$ for simplicity. 
Then the dynamic regret is
\begin{align*}
	\text{Reg}_T = \sum_{t=1}^{T}\max_a\mu_{a,t} - \mathbb{E}\left[\sum_{t=1}^{T}\mu_{A_t,t}\right].
\end{align*}
Unlike the regret definition for the adversarial setting that is with respect to the best single arm, dynamic regret is defined with respect to the sequence of best arms at each time step.
There are two popular constraints that quantify the reward changes.
One is the total count of changes in the mean reward that occur before time $T$:
\begin{align*}
	S_T = 1+\sum_{t=2}^{T}\mathbb{1}{\{\mu_{a,t-1}\neq\mu_{a,t}, \text{ for some $a$}\}}.
\end{align*}
Settings where the above quantity is assumed to be small are also called piece-wise stationary.
Another constraint quantifies the total variation for the reward mean:
\begin{align*}
	B_T = \sum_{t=2}^{T}\max_a |\mu_{a,t} - \mu_{a,t-1}|.
\end{align*}
\paragraph{Piecewise Stationary Setting.}
The first algorithm designed for the non-stationary MAB under finite number of reward mean changes is Exp3.S~\citep{auer2002nonstochastic}, which is a variant of the Exp3 algorithm.
The expected regret of Exp3.S is $O(S_T\sqrt{KT\log(KT)})$ in general, but can be improved to $O(\sqrt{S_TKT\log(KT)})$ if $S_T$ is used to tune the algorithm parameters.
Later on, \citet{garivier2011upper} proved that no policy can achieve problem-dependent regret smaller than $O(\sqrt{T})$ in the non-stationary case, and Exp3.S is thus optimal up to logarithmic factors of $T$.
\citet{garivier2011upper} also studied two UCB-based algorithms: discounted UCB (D-UCB) which was first proposed by~\citet{kocsis2006discounted} and sliding-window UCB (SW-UCB). 
Their main idea is to encourage using more recent data in UCB reward estimations. 
When $S$ is given, both algorithms achieve $O(K\sqrt{ST}\log T)$ regret. 
Another body of works explore the idea of monitoring the reward distributions by change-detection methods and reset the bandit algorithm accordingly. 
Using this idea, Monitored-UCB (M-UCB)~\citep{cao2019nearly} combines the UCB algorithm with a change-point detection component based on running sample means over a sliding window and achieves $O(\sqrt{KST\log T})$ regret under certain assumption on the reward mean change. 
In same year, \citet{auer2019adaptively} proposed an action elimination algorithm ADSWITCH that again detects changes in the mean reward and restarts the learning algorithm accordingly. 
Without knowing $S$ nor making other assumptions, ADSWITCH achieves $O(\sqrt{KST\log T})$ regret. 

\paragraph{Bounded Total Variation Setting.}
Comparing with piecewise stationary setting, bounded total variation is a softer constraint. 
Here nature has the power to change the reward at every round but only up to a total amount limit $B_T$. 
\citet{besbes2014stochastic} proposed the Rexp3 policy that uses the famous Exp3 algorithm as a subroutine and restarts it at every batch. 
Rexp3 with a batch size tuned by $B_T$ has $O((B_TK\log K)^{1/3}T^{2/3})$ regret that nearly matches the lower bound of $\Omega(KB_T)^{1/3}T^{2/3})$ in this setting~\citep{besbes2014stochastic}.

The non-stationary setup has also studied in linear bandits and contextual bandits. 
We briefly describe the latter one as an example.
Define the distribution of the context-reward pairs $\mathcal{C}\times[0,1]^K$ by $\mathcal{D}_1,\ldots,\mathcal{D}_T$. 
At each round, the environment samples $(c_t,x_t)$ and reveals $c_t$ to the agent, then the agent picks an arm $A_t\in[K]$ and observes $x_t(A_t)$. 
For a fixed set of policies $\Pi$ that contains mappings: $\mathcal{C}\rightarrow[K]$, the dynamic regret is defined as:
\begin{align*}
    \text{Reg}_T = \sum_{t=1}^T\max_{\pi\in\Pi}\mathbb{E}_{(c,x)\sim\mathcal{D}_t}[x(\pi(c))] - \sum_{t=1}^Tx_t(A_t).
\end{align*}
Similar to the MAB setting, there are two ways to measure the non-stationary of the environment: total number of changes and the total variation in below.
\begin{align*}
    S &= 1+\sum_{t=2}^T\mathbb{1}\{\mathcal{D}_t\neq\mathcal{D}_{t-1}\},\\
    B_T &= \sum_{t=2}^T\left\|\mathcal{D}_t-\mathcal{D}_{t-1}\right\|_{TV}.
\end{align*}
\citet{chen2019new} proposed $\text{ADA-ILTCB}^+$ algorithm that is parameter-free, efficient and achieves optimal regret $\widetilde{O}(\min\{\sqrt{K(\log |\Pi|)ST}, \sqrt{K(\log|\Pi|)T}+(K\log |\Pi|B_T)^{1/3}T^{2/3}\})$ by randomly entering replay phases to detect non-stationarity. More recently, a generic reduction has been studied that allows one to convert certain algorithms for the stationary setting into algorithms for the non-stationary setting \citep{wei2021non}. What is nice about this work is that the resulting algorithms often \emph{simultaneously} achieve optimal guarantees in terms of both $S$ and $B_T$ \emph{without} prior knowledge of any of these parameters.

\subsection{Robustness}\label{sec:robust}
So far, we have discussed stochastic and adversarial bandits.
Stochastic bandit is an ideal setting, since it assumes a fixed reward distribution. Adversarial, on the other hand, considers a too extreme scenario. In real-world applications, some of the observed rewards may be corrupted resulting in deviation from purely stochastic behavior. For instance, there might be clerical errors while creating electronic health records or sensor errors in recording health status from wearable devices. If the reward is self-reported by patients, there can be corruptions due to mistakes, lack of attention, boredom, etc. In this section, we consider algorithms that are \textit{robust} to possible corruptions. Note that this setting is between purely stochastic and adversarial but in a different way than the  non-stationary bandit discussed in Section \ref{sec:non_stationary}. 

\paragraph{Fraction corruption model.} The fraction corruption model limits the fraction of the total number of rounds that an adversarial corruption can happen.
Assume corruptions happen with a probability $\eta \in [0, 1]$. The adversary closely follows the progress of 
arm pulling and reward generation. At each time step $t$, after the algorithm has decided to pull an arm $A_t$, the adversary first decides whether to corrupt this arm pull or not by performing a Bernoulli trail $Z_t \in \{0, 1\}$ with a mean $\eta \in [0, 1]$.
Then it generates a corruption $\zeta$ arbitrarily. After this, the ``clean reward'' $R_t^*$ is generated by the environment. The reward a player receives at  step $t$ is 
$$
    R_t = \mathbb{1}\{Z_t = 0\} R_t^* + \mathbb{1}\{Z_t = 1\} \zeta_t. 
$$

Since the rewards may be contaminated, directly running the algorithms built for stochastic bandits can result in degraded performance. Several papers have proposed variants of UCB that use a more robust estimate. \cite{kapoor2019corruption} proposed RUCB-TUNE (Robust UCB) which makes two crucial changes to the classical UCB:  RUCB-TUNE uses the median and a tuned variance estimate instead of the simple sample mean and variance to construct its upper bound. The algorithm requires an upper bound $\eta_0$ on the corruption rate $\eta$. Their analyses are built on the environment with Gaussian reward distributions and they have a regret bound of $\mathcal{O}((1-\eta)\sqrt{KT \log(T)} + \eta_0(\mu^* + B)T)$, where $B$ is an upper bound on the scales of corruptions $|\zeta_t|$ and $\mu^*$ is the maximum expected reward of all the arms. They provide another algorithm RUCB with a similar result. RUCB does not require the knowledge of $\eta_0$ but it requires an upper bound on the variance of rewards for all the arms.

\cite{niss2020you} considers a stronger requirement that all the arms have no more than $\eta$-fraction of corruptions. They proposed crUCB (contamination robust-UCB), which mimics RUCB in the general framework. The only difference is that they allow a variety of robust mean estimates including $\alpha$-trimmed mean, mean estimates that ignores the largest $(1-\alpha)$-fraction of data, $\alpha$-shorth mean. Their algorithms are evaluated by uncontaminated regrets, the regret between true rewards of the best possible actions and the algorithm-chosen actions:
$$
    \text{Reg}_T = \max_{a\in\mathcal{A}} \mathbb{E}[\sum_{t = 1}^T R^*_a(t) - R^*_{A_t}(t)],
$$
If the contamination fraction is small enough, both crUCB using $\alpha$-shorth mean and $\alpha$-trimmed mean achieve uncontaminated regret of $ \sqrt{K T \log (T)}+\sum_a \Delta_{a}$, where $\Delta_a$ is the gap between the expected reward of $a$ and the optimal action.

 \paragraph{Budget-bounded corruption model.} In budget-bounded corruption model \citep{lykouris2018stochastic}, all the rounds can be possibly corrupted but the total deviation between the ``clean rewards'' and the corrupted rewards is bounded. We define the total corruption budget as 
 $$
    B_T = \sum_{t = 1}^T \|R_t^* - R_t\|. 
 $$
 \cite{gupta2019better} proposed BARBAR (Bandit Algorithm with Robustness: Bad Arms get Recourse) based on the arm elimination algorithm discussed in Section \ref{sec:basics} with some crucial modifications that make it robust to corruptions. \cite{gupta2019better} show that their algorithm achieves a gap-dependent regret bound of
 $$
    \text{Reg}_T = \mathcal{O}\left(K B_T+\sum_{a \neq a^{\star}} \frac{\log T}{\Delta_{a}} \log \left(\frac{K}{\delta} \log T\right)\right).
 $$
 \emph{without} knowing the value of $C$.

\subsection{Dealing with Constraints}\label{sec:safety}
In standard bandit settings, the only criterion to evaluate the performance of an algorithm is the regret (either cumulative or simple regret). 
In this section, we introduce constrained bandit problems, where the agent also needs to satisfy certain constraints while minimizing its regret. 
Depending on the actual problem's demands, different types of constraints have been considered. 
For healthcare applications, two types of constraints frequently arise in practice. 
First, delivering an intervention might consume resources such as doctor's time, patient's attention, phone battery and power, etc. We might therefore want to maximize rewards subject to a budget on resource consumption. 
Second, we might be worried that uncontrolled exploration might yield performance that is substantially worse than an existing standard of care. 
With these concerns in mind, in this section we consider the following two settings: bandits with knapsacks and conservative bandits.

\paragraph{Bandits with Knapsacks (BwK). }
The name of \emph{bandits with knapsacks} comes from the well-known knapsack problem in combinatorial optimization that studies packing items into a fixed-size knapsack.
In a knapsack problem, each item has its value and size.
Its ultimate goal is to fill the knapsack with as large value as possible.
The \emph{bandits with knapsacks} then focus on a stochastic online version of the knapsack problem.
Many practical scenarios can be captured by this framework.
For example, in medical trials, physicians may be limited by the cost of treatment materials while optimizing the health condition of the patients.
In online recommendations, the website designer may be constrained by the advertisers' budgets while maximizing its profit.
We follow the terminologies and notations in~\citet{badanidiyuru2018bandits} to formally describe this setting and their proposed algorithms.

An agent is given an action set $\mathcal{A}$ and there are $d$ resources being consumed.
At every round, the agent selects an action $A_t\in\mathcal{A}$ and observes a reward $X_t$ along with a $d$-dimensional resource consumption vector $\mathbf{c}_t$.
For each resource $i\in[d]$, the total consumption should not exceed the pre-specified budget $B_i$ at any round. 
The agent stops immediately when the total consumption of certain resource exceeds its budget. 
Denote the stopping time by $\tau$, the goal is to maximize the total reward until time $\tau$.
The worst-case regret for an algorithm is defined as the difference between the benchmark total reward $\mathrm{OPT}$\footnote{An optimal dynamic policy that maximizes the expected total reward given prior knowledge on all latent distributions such as the reward distribution and the cost consumption for each action.} and the algorithm's expected total reward until $\tau$.

To solve the problem, \citet{badanidiyuru2018bandits} proposed two algorithms.
The first one is a primal-dual algorithm called $\mathrm{PrimalDualBwK}$. 
At every round, it estimates the upper confidence bound for the expected reward and the lower confidence bound for the resource consumption for each arm. 
Then $\mathrm{PrimalDualBwK}$ plays the most ``cost-effective'' arm, i.e., the one with highest ratio of the expected reward's upper bound to the expected cost. 
The regret of $\mathrm{PrimalDualBwK}$ is proved to be $\widetilde{O}\left(\sqrt{|\mathcal{A}|\mathrm{OPT}}+\mathrm{OPT}\sqrt{|\mathcal{A}|/B}\right)$, where $B:=\min_i B_i$.
Note that without the resource constraints, the regret bound becomes $\widetilde{O}\left(\sqrt{|\mathcal{A}|\mathrm{OPT}}\right) = \widetilde{O}\left(\sqrt{|\mathcal{A}|T}\right)$ that is optimal up to logarithmic factors since the only constraint is the time horizon $T$ and we can simply set $B = T$ in the general regret formula.
The second algorithm is called $\mathrm{BalancedExploration}$. 
Its design principle is to explore as much as possible while avoiding obviously sub-optimal strategies.
They show that the regret of $\mathrm{BalancedExploration}$ is $\widetilde{O}\left(\sqrt{d|\mathcal{A}|\mathrm{OPT}}+\mathrm{OPT}\sqrt{d|\mathcal{A}|/B}\right)$. 
Even though its regret has worse dependence on $d$ comparing with $\mathrm{PrimalDualBwK}$, $\mathrm{BalancedExploration}$ performs better in some special cases. 

Comparing with the general constraint in above knapsack framework, researchers have also considered more specific settings.
For example, other than the time horizon, there is a single resource with deterministic consumption and different arms consume the resource at different rates~\citep{tran2010epsilon,tran2012knapsack,ding2013multi,xia2015thompson,zhou2018budget}. 
Such problems are often called \emph{budgeted bandits}.
\citet{agrawal2014bandits} generalized the BwK model by allowing arbitrary concave rewards and convex constraints.
Furthermore, similar constrained bandit problems are also studied in settings that includes contextual bandits~\citep{agrawal2014bandits,wu2015algorithms,agrawal2016linear} and even adversarial bandits~\citep{sun2017safety,immorlica2019adversarial}.

\paragraph{Conservative Bandits.}
It is well-known that standard bandit algorithms often explore wildly in their early stages, so their regret over initial rounds can be very high.
However, safety is one of the crucial concerns for designing adaptive experiments, so the early stage high regret cannot be tolerated in some high stakes scenarios.
For instance, the UCB algorithm can assign treatments almost randomly to a patient during the first several rounds, which may cause severe problems for the patient's overall health.
A company may also not be able to withstand the extremely low revenue initially if its operation is in great need of cash flow.
In these examples, the real situation is that the physician or the company already have their favorite policies that operate well. 
They would like to explore new strategies to optimize their treatment performance or revenue while maintaining their performance to not be significantly worse than a baseline, uniformly over time.
\emph{Conservative bandits}~\citep{wu2016conservative} exactly model this problem.
We follow their terminology and notation to describe the problem framework.

We use $\{0, 1,\ldots, K\}$ as the indices of actions, in which the arm indexed by $0$ correspond to the default action (the agent's typical strategy) and the other arms are the alternatives to be explored.
The agent selects an arm $A_t\in\{0,1,\ldots,K\}$ in round $t$.
Denote $R_{t,i}$ as the random reward received at time $t$ after playing arm $i$, then the regret and pseudo-regret are defined as $\text{Reg}_T = \max_{i\in\{0,1,\ldots,K\}} \sum_{t=1}^TR_{t,i} - R_{t, A_t}$ and $\widetilde{\text{Reg}}_T = \max_{i\in\{0,1,\ldots,K\}}\sum_{t=1}^T \mu_i - \mu_{A_t}$.
To ensure that the agent performs as good as his or her usual strategy, the algorithm needs to satisfy:
\begin{align}
    \sum_{s=1}^t R_{s, A_s} \geq (1-\alpha)\sum_{s=1}^t R_{s,0}, 
    \label{equ:conservative_constraint}
\end{align}
where $0<\alpha\leq 1$.
Above constraint guarantees that the reward collected by the agent is at least $(1-\alpha)$ fraction of the reward from simply playing arm $0$.
The objective of conservative bandits algorithms is to minimize regret while satisfying \eqref{equ:conservative_constraint} for all $t$.
To solve the problem, ~\citep{wu2016conservative} proposed a novel algorithm called conservative UCB which is based on UCB with the novel twist of maintaining \eqref{equ:conservative_constraint} being satisfied.
Basically, the agent follows the UCB suggestion if \eqref{equ:conservative_constraint} can be satisfied from their estimation, otherwise, the agent switches to the conservative arm $0$.
They show that conservative UCB achieves pseudo-regret $\widetilde{\text{Reg}}_T = \widetilde{O}(\sqrt{KT}+\frac{K}{\alpha\mu_0})$, where $\mu_0$ denotes the expected reward for arm $0$.
For more details, including gap-dependent regret and regret analysis for the adversarial setting, we refer the reader to~\citet{wu2016conservative}.

The concept of conservative bandits, namely performing as good as a baseline uniformly over time, is quite general. 
Recently, it has been generalized to more settings, e.g., conservative contextual bandits~\citep{kazerouni2016conservative,garcelon2020improved} and conservative reinforcement learning~\citep{garcelon2020conservative}.

\subsection{Fairness}\label{sec:fair}
Algorithmic fairness has become an increasingly important topic in machine learning research \citep{barocas-hardt-narayanan}. Fairness is also a concern in healthcare applications due to the biases in data collection and algorithmic design \citep{paulus2020predictably}. Using healthcare cost as a proxy for healthcare needs leads to biased risk scores that hurt Black patients \citep{obermeyer2019dissecting}. Using genetic datasets collected mostly from patients of European ancestry leads to biased genetic risk scores on patients with non-European ancestries \citep{martin2019clinical}. Unfairness can also be rooted in algorithmic design. Algorithms that simply maximizing user responses can be unfair in how they allocate exposure to the items in recommendation systems \citep{singh2018fairness}. In this section, we review several popular definitions of fairness in bandit literature and discuss how to calibrate unfairness.


\paragraph{Two types of fairness.} Though there remains little agreement about what “fairness” should mean in different contexts, the literature can be divided into two broad families: those that target \textit{group} fairness and those that target \textit{individual} fairness. Group fairness requires the algorithm to maintain fairness across different demographic groups (say by race or gender) that are supposed to be treated equally. As we mentioned above, biased healthcare algorithms may hurt patients from certain demographic groups due to low representative. Individual fairness, on the other hand, asks for some constraints on the individual level, including two slightly different types: fairness through awareness, which requires that similar individuals be treated similarly and meritocratic fairness, which requires that less qualified individuals not be favored over more qualified individuals. Individual fairness in healthcare is also important: for two patients with similar demographic information or physical condition, we shall not develop an algorithm that performs much better for one patient while useless (or even harmful) for the other. 

In the context of bandits, the literature can be further divided into those that model individuals as arms in the MAB setting and those that models individuals as contexts in the contextual bandits setting. We give some examples of fairness definitions from each category.

\paragraph{Individual fairness.} We start from individual fairness in the MAB setting with each arm representing an individual. \cite{joseph2016fairness} proposed the use of meritocratic fairness. Formally, an algorithm is said to be $\delta$-fair if over $T$ time steps, , for any pair of arms $a$, $a'$, and any round $t$,
$$
    \pi_t(a) > \pi_t(a') \text{ only if } \mu_a > \mu_{a'}
$$
with probability at least $1-\delta$. 
This requires the algorithm to give equal probability to choose two arms unless one there is strong evidence that $\mu_a > \mu_{a'}$. \cite{joseph2016fairness} developed an algorithm that is $\delta$-fair based on the traditional UCB algorithm. In specific, they maintain a set of arms that chains to the arm with highest UCB. All the arms in the set are given the same probability to be selected. Their regret bound is $\mathcal{O}(\sqrt{K^{3} T \ln \frac{T K}{\delta}})$, with a higher cubic dependence on $K$, which is the price they pay to give equal probability to all the arms in the set. More generally, \cite{liu2017calibrated} considers a fairness through awareness. An algorithm is said to be $(\epsilon_1, \epsilon_2, \delta)$-smooth fair \footnote{Note that \cite{liu2017calibrated} considers a general divergence function for the distribution of action selection and rewards, we introduce a special case here for a cleaner form.}, if for any pair of arms $a$, $a'$ and any $t$, with a probability at least $1-\delta$,
$$
    |\pi_t(a) - \pi_t(a')| \leq \epsilon_{1} |\mu_a -  \mu_{a'}|+\epsilon_{2}.
$$
They introduced a fairness regret to quantify the violation that occurs when the arm with the highest reward realization at a given time is not selected with the highest probability. Formally, they have 
$$
    \mathrm{Reg}_T = \sum_{t = 1}^T\mathbb{E}\left[\sum_{i = 1}^K \max\{P^*(i) - \pi_t(i), 0\}\right], 
$$
where $P^*(i)$ is the probability that arm $i$ has the highest regret. 
They developed a modified Thompson Sampling algorithm with an initial uniform exploration phase that achieve fairness regret of $\mathcal{O}((KT)^{2/3})$. \cite{wang2021fairness} also aims at the fairness through awareness by comparing to the optimal fair policy $\pi^*$,
$$
    \frac{\pi^*(a)}{\pi^*(a')} = \frac{f(\mu_a)}{f(\mu_{a'})} \text{ for some metric function } f.
$$ 
They in turn target at the fairness regret, the cumulative $L_0$ norm of $\pi^* - \pi_t$.
Other individual fairness definition that simply requires a minimum number of pulls for each arms is considered by \cite{patil2020achieving} and \citet{chen2020fair}.

\paragraph{Group fairness.} For group fairness, one may divide arms into $n$ groups $G_1, \dots, G_n$. \cite{schumann2019group} proposed an algorithm ensuring that the probability of pulling an arm does not change based on group membership: 
$$
    \operatorname{P}(\text{pull } a \mid a \in G_i) = \operatorname{P}(\text{pull } a \mid a \in G_j), \forall i, j < n \text{ and } a \in \mathcal{A}.
$$ In other words, one cannot prefer one arm over the other one based on the group information.

Some literature treats each context $c_t$ as an individual and aims at the fairness among different contexts.
\citep{huang2020achieving} divides the context space into two groups, a privileged group $G^+$ and a protected group $G^{-}$. A group-level cumulative mean reward is defined as 
$$
    \bar{R}^{G} = \frac{1}{|T_G|} \sum_{t \in T_G} R_{t}, \text{ where } T_G \text{ is the set of all rounds with $g_t$ in group $G$.} 
$$
Their group fairness requires that $|\mathbb{E}[\bar{R}^{G^+} - \bar{R}^{G^-}]| \leq \tau$, where $\tau \in \mathbb{R}^+$ reflects the tolerance degree of unfairness. Assuming a linear bandit setting, they proposed a Fair-LinUCB algorithm, that penalize the arms that are unfair by decreasing the corresponding UCB values. 

\paragraph{Counterfactual fairness.} Another line of work \citep{kusner2017counterfactual} models fairness through causal inference. Their fairness definition is normally referred to as Counterfactual Fairness. In the group fairness, we require the distributions of the value of interest are the same across different demographic groups. However, what we really want is that the demographic group information does not cause the unfairness, which can not be fully inferred from the group fairness definition. Counterfactual fairness, on the other hand, requires the distribution of the value of interest does not change when the demographic group were changed from one to the other while keeping all the other context variable unchanged.

\paragraph{General fairness metric.} All the above methods requires certain fairness definition, it is natural to ask whether there is an universal algorithm that adapts to various fairness definition. In real-applications, we may not know the exact form of fairness metric. Instead, we may know whether certain policy violates fairness condition.  Instead of studying a specific fairness metric, \cite{gillen2018online} developed an algorithm that achieves fairness through a fairness oracle, which tells the algorithm whether fairness is violated. This also allows great flexibility in the choice of fairness metric. At each round, each arm $i \in [K]$ is given a context $C_{t, i} \in \mathbb{R}^d$. The probability of selecting action $i$ at the step $t$ is denoted by $\pi_{t, i}$, which is a vector denoting the probability of selecting . We denote the vector of context (policy) at round $t$ by $c_t$ ($\pi_t$). The fairness oracle is define as following. Intuitively, it tells the algorithm, which pairs of actions violate the fairness metric. 
\begin{defini}
Let $\Delta([K])$ be the set of all distributions over $[K]$. Given a user-specified distance function $d$, a fairness oracle $O_d$ is a function $O_d: \mathbb{R}^{d \times K} \times \Delta([K]) \rightarrow 2^{[K] \times[K]}$, defined such that
$$
    {O}_{d}\left(c_{t}, \pi_{t}\right)=\left\{(i, j):\left|\pi_{t, i}-\pi_{t, j}\right|>d\left(C_{t, i}, C_{t, j}\right)\right\}.
$$
\end{defini}
They developed an algorithm that only has access to the fairness oracle. The algorithm has the number of fairness violations depending only logarithmically on $T$ and a regret bound of the optimal rate $\mathcal{O}(\sqrt{T})$ with respect to the best fair policy.

\subsection{Benefiting from Causal Knowledge}\label{sec:causal}
Causal bandit is an example of a structured bandit problem where actions are composed of interventions on variables of a causal graph. 
According to the underlying causal dynamics, performing an intervention can help learn another intervention's reward distribution.
Algorithms for causal bandit problems exploit the causal dependency among interventions to reduce the regret or sample complexity.

\cite{bareinboim2015bandits} was the first to connect bandit problems with causal models.
One of the main issues that make causal inference hard is the existence of confounders defined below, which can confuse correlation and causation.
\cite{bareinboim2015bandits} pointed out the importance of capturing the confounders before actually performing actions.
In particular, they proposed a new criterion for choosing actions called regret decision criterion (RDC).
Under RDC rule, the agent collects her \emph{intention} (includes information on confounders) on selecting actions and re-thinks based on the estimated reward statistics for the current \emph{intention} and then decides which action to play.
Since the actual selected action can be different than the agent's \emph{intention}, the agent decides in a counterfactual way.
\begin{defini}[Confounder]
A confounder is a variable that influences both the dependent variable and independent variable.
\end{defini}

Later on, \citet{lattimore2016causal} formally proposed the causal bandit framework via causal graphs.
A causal model consists of a directed acyclic graph $G$ over a set of random variables $\mathcal{V} = \{V_1,\ldots, V_n\}$ and a joint distribution $P$ that factorizes over $G$.
A size $m$ hard intervention (action) is denoted by $\text{do}(\mathbf{V}=\mathbf{v})$, which assigns the values $\mathbf{v} = \{v_1,\ldots, v_m\}\subset\mathcal{V}$ to the corresponding variables $\mathbf{V} = \{V_1,\ldots, V_m\}$. 
In causal bandits, the action set is defined as
\begin{align*}
    \mathcal{A}:=\{\text{do}(\mathbf{V}=\mathbf{v})|\mathbf{V}\subset\mathcal{V}, \mathbf{v}\in\text{Dom}(\mathbf{V})\},
\end{align*}
or its subset and the optimal intervention is $a^*:=\argmax_{a\in\mathcal{A}}\mathbb{E}[R|a]$.

Many practical problems can be modeled via above framework. 
For example, in healthcare applications, the physician adjusts several features such as dose levels on different medicines or life-style advices to achieve some desirable clinical outcomes~\citep{liu2020reinforcement}.
Genetic engineering also involves direct manipulation of genes using biotechnology to produce improved organisms. 
In these problems, the number of interventions can be exponentially large in the number of manipulable variables so that standard MAB algorithms cannot work efficiently.
To address this issue, recent work on causal bandits has developed methods that exploit causal information to achieve simple or cumulative regret that does not scale with the number of interventions~\citep{lattimore2016causal, sen2017identifying, lee2018structural,lu2020regret,lu2021causal}.
We summarize existing methodologies below.

\paragraph{Refining the Policy Space.}
Assuming the causal graph is known (confounders may also exist), \citet{lee2018structural} proposed an intervention set reduction algorithm.
They showed that not all interventions are worthwhile to be played, because there are equivalences among interventions and some interventions can be proved to be no better than others. 
Their algorithm filters out redundant interventions using the causal graph structure before applying any standard MAB algorithm on the reduced intervention set, also called \emph{possibly-optimal minimal intervention set} (POMIS).
Since the size of POMIS is usually much smaller than $|\mathcal{A}|$, the corresponding regret can be reduced. 
Follow-up work in~\citet{lee2019structural} further showed that the expected reward for interventions in a POMIS can be estimated from each other in certain cases.
Their method also works well when some variables are non-manipulable.
We remark that the step of finding the POMIS solely relies on the given causal graph structure and does not involve performing any intervention. 
\paragraph{Accelerating Using Causal Graph Side Information in Standard MAB Algorithms.}
Recall that a key step in bandit algorithm is estimating the expected reward for each arm $\hat{\mu}_a$. 
Intuitively speaking, one needs to play each arm for enough number of times in order to get accurate estimations if interventions are independent. 
But in causal bandits, interventions are correlated with each other so that performing one intervention can help with estimating others according to the shared causal model. 
\citet{lattimore2016causal} studied the best arm identification problem via importance weighting.
Their algorithm achieves simple regret $O(\sqrt{m/T})$, where $m$ can be smaller than $|\mathcal{A}|$ in many scenarios.
\citet{sen2017identifying} further generalized the algorithm in \citet{lattimore2016causal} to a soft intervention setting and proved gap-dependent regret guarantees.
In soft interventions, the agent does not directly set values to specific variables, but instead change their conditional probabilities given their direct parents.
\citet{lu2020regret} proposed causal UCB and causal TS algorithms that achieve $\tilde{O}(\sqrt{ZT})$ cumulative regret, where $Z$ is a graph dependent number and can be exponentially smaller than $|\mathcal{A}|$.
A causal linear bandit framework and its efficient algorithms were also well studied.
Furthermore, gap-dependent worst-case regret bound and a budgeted causal bandit setting was also considered in follow-up works~\citep{nair2021budgeted}.

\subsection{Multi-task Learning}\label{sec:multi_task}
Multitask learning is the learning paradigm in machine learning that aims at leveraging information in multiple tasks to achieve a better sample efficiency by solving them jointly rather than separately \citep{zhang2021survey}.
Multitask learning algorithms usually assume some similarity between tasks that allows information sharing. In the context of multitask bandit, each bandit is called a task and various similarity assumptions on the reward distribution can be made. This can model many interesting scenarios. For instance, a group of patient with the same disease can be seen as multiple tasks. We can not directly use the same policy, as a personalized treatment should be applied to each patient. However, some patients' responses may be similar to the other's such that one can transfer the information across patients. Recall that we use $\nu = (P_{a}, a \in \mathcal{A})$ denote a bandit instance. A general multitask bandit framework can be summarized below:
$$
    \text{ Given a set of bandits } \nu_1, \dots, \nu_n,\ \operatorname{Discrepancy}(\nu_i, \nu_j) \text{ is small for some } i, j \in [n],
$$
for some discrepancy measure on bandit instances.

In this subsection, we will review some interesting works on multitask bandit and compare their assumptions on discrepancy and the corresponding regret bounds. We will also cover the potential applications of multitask bandits.

\paragraph{Parametric Model.} Early work on multitask bandit was done by \cite{cesa2013gang} who considered a gang of contextual bandits connected with a graph $\{V, E\}$, where $V = \{1, \dots, n\}$ is the node set and $E$ is the edge set.  Each node $i$ is assumed to be a linear bandit with rewards generated by $R_i(c) = \theta_i^Tc + \epsilon_i(c)$ given a context $c$, where $\epsilon_i(c)$ is a conditionally zero-mean noise. Any two nodes that are adjacent on the graph have low discrepancy compared to the total scale:
$\sum_{i \in V}\|\theta_{i}\|^{2} \gg \sum_{(i, j) \in E}\|\theta_{i} - \theta_j\|^{2}.$ Their discrepancy measure is simply the $L_2$ distance on the unknown parameters. However, their analysis only shows a benefit of a factor $n$ in the logarithmic term. In the literature of supervised multitask learning, the benefit could be a factor of $n^{1/2}$ outside the logarithmic term \citep{maurer2016benefit}. Another property makes it hard for their algorithm to generalize to other settings is that the discrepancy information on tasks similarities is partially known to the agent through the graph structure. Generally speaking, the agent does not know which tasks are similar and have to adaptively learn the information. 

As an improvement, \cite{gentile2014online,li2016collaborative} consider an \textit{unknown} graph with clustered structure, and all the bandits within the same cluster share a common reward distribution, which is also from a linear model. The algorithm is to simply cluster the bandits based on the feedbacks and pool the data from all the bandits from the same cluster. Instead of have a dependency of $n$ as for the baseline algorithm that learns all the bandit separately, their regret is $\tilde{\mathcal{O}}({dm\sqrt{T}})$, where $m \ll n$ is the number of clusters.

\paragraph{Non-parametric Model.} Despite the encouraging results by \cite{gentile2014online,li2016collaborative}, the clustered structure and linear payoffs are strong assumptions. \cite{deshmukh2017multi} proposed a kernel regression model that allows general similarities. Formally, they assume the reward is generated by mean function $f:\mathcal{Z}\times \mathcal{C} \mapsto \mathbb{R}$, where $\mathcal{Z}$ is called the task similarity space, $\mathcal{C}$ is the context from the context space $\mathcal{C}$. Let $\tilde{k}$ be a SPD (semi-positive definite) kernel on $\mathcal{Z} \times \mathcal{C}$ with the form
$$
    \tilde{k}\left((z, c),\left(z^{\prime}, c^{\prime}\right)\right)=k_{\mathcal{Z}}\left(z, z^{\prime}\right) k_{\mathcal{C}}\left(c, c^{\prime}\right),
$$
where $k_{\mathcal{Z}}$ and $k_{\mathcal{C}}$ are SPD kernels on $\mathcal{Z} = \{1, \dots, n\}$ representing $n$ tasks and $\mathcal{C}$. They assume that $f$ is from the RKHS, $\mathcal{H}_{\tilde{k}}$ corresponding to $\tilde{k}$. Kernel $k_{\mathcal{Z}}$ characterized the similarity between different bandits, for instance:
\begin{enumerate}
    \item Independent: $k_{\mathcal{Z}}(z, z') = \mathbb{1}_{z = z'}$, i.e. observations from different tasks are not sharing any information.
    \item Pooled: $k_{\mathcal{Z}} \equiv 1$, i.e all the observations from different tasks are treated the same.
    \item Multi-task: $k_{\mathcal{Z}}$ is a PDS matrix reflecting task similarity.
\end{enumerate}

Their algorithm \textbf{KMTL-UCB} gives an regret bound of $\tilde{\mathcal{O}}(\sqrt{Tr_z r_c})$, where $r_z$ is the rank of the $n\times n$ matrix $K_{Z}=\left[k_{\mathcal{Z}}\left(z_{i}, z_{i}\right)\right]_{i=1}^{n}$ and $r_c$ is the rank of $K_{C_T} \coloneqq [k_{\mathcal{C}}(c_{a_t, t}, c_{a_{t'}, t'})]_{t, t' = 1}^T$. To compare, the regular regret bound of learning one task independently is $\tilde{\mathcal{O}}(\sqrt{Tr_c})$ and all tasks independently gives $\tilde{\mathcal{O}}(n\sqrt{Tr_c})$. The scaling of $r_z$ determines the benefits of multitask learning: $r_z = n$ corresponds to the first case above with $k_{\mathcal{Z}}(z, z') = \mathbb{1}_{z = z'}$ and $r_z = 1$ corresponds to the second case with $k_{\mathcal{Z}} \equiv 1$. Though \cite{deshmukh2017multi} gives a general way to characterize the multitask bandit benefits, their algorithm still relies on the knowledge of the similarity kernel $k_{\mathcal{Z}}$. In practice, one can learn a kernel, but little theory has been established to justify this practice.  


\paragraph{Adaptive Learning.} \cite{tomkins2021intelligentpooling} proposed Intelligent Pooling, a generalization of a Thompson sampling contextual bandit for learning personalized treatment policies. They assume the reward for user $i \in [n]$ is generated by
$$
    R_i = \phi(C_i, A_i)^T \theta_i +\epsilon_i,
$$
where $\phi: \mathcal{C}\times \mathcal{A} \mapsto \mathbb{R}^d$ is the pre-specified mapping from contexts and actions and $\theta_i \in \mathbb{R}^d$ is the personalized parameter to learn. IntelligentPooling imposes each $w_i$ as a random-effects upon a population-level parameter $w_{pop}$, i.e. $w_i = w_{pop} + u_i$ with $u_i$ the Gaussian random effect. Prior distribution of $w_{pop}$ is also assumed to be Gaussian. IntelligentPooling runs by sampling $w_{i, t}$ from the posterior distribution of $w_i$. The algorithm uses random effects to adaptively pool users' data based on the degree to which users exhibit heterogeneous rewards that is away from the population behavior ($w_{pop}$). They showed a regret bound of $\tilde{O}(dn\sqrt{T})$.

\section{Conclusion}\label{sec:conclusion}

We reviewed basic topics in bandit algorithms that are helpful in understanding their increasing role in mobile health, and more broadly, precision medicine. We also discussed a selection of advanced topics that might inform the design of the next generation of bandit driven mobile health applications. We hope that the reader has gained an appreciation of the beauty and power of bandit algorithms and of their relevance to mobile health and precision medicine. However, like any methodology, bandit algorithms do have their limitations. First, bandit algorithms assume that an appropriate reward is available upfront. In practice, especially in healthcare applications, the design of rewards to be optimized is a difficult problem with no easy solution. The issues involved in specified a reward function that is aligned with the goals of various human stakeholders have been brought into focus more recently under the term ''the alignment problem'' \citep{christian2020alignment}.
Second, bandit algorithms do not take into account delayed impacts of the agent's actions. When actions have delayed impacts, it might be beneficial to take an action with low immediate reward if it increases the probability of higher rewards later on. The full treatment of this problem of delayed impact of actions requires techniques from reinforcement learning \citep{sutton2018reinforcement}. In fact, bandit problems can be thought of as the simplest of reinforcement learning problems.

\bibliographystyle{apalike}
\bibliography{ref}
\end{document}